\documentclass[10pt,twocolumn,letterpaper]{article}
\usepackage[usenames, dvipsnames]{xcolor}
\usepackage{usenix2019, times,color,graphicx,amsmath,array,amssymb, subcaption}
\usepackage{mathptm,times,helvet,courier,xspace}
\usepackage{boxedminipage,multirow,endnotes,balance}
\usepackage{pdfpages}
\usepackage{rotating}
\usepackage{colortbl}
\usepackage{transparent}
\usepackage{appendix}
\usepackage{multirow}
\usepackage{relsize}
\usepackage{booktabs}
\usepackage{multirow}
\usepackage{cite}

\usepackage{cancel}

\usepackage{listings}

\usepackage[]{hyperref}
\usepackage{xurl}

\newcommand{\eg}{{e.g.,} }

\newcommand{\nop}[1]{}

\newcommand{\Sec}[1]{$\S$\ref{s:#1}\xspace}

\newcommand{\tightcaption}[1]{\vspace{-11pt}\caption{{\bf \small #1}}
\vspace{-15pt}
}
\newcommand{\para}[1]{\smallskip\noindent{\bf #1}}

\newcommand{\cut}[1]{}

\newcommand{\name}{Boggart\xspace}
\def\compactify{\itemsep=0pt \topsep=0pt \partopsep=0pt \parsep=0pt}
\let\latexusecounter=\usecounter

\newenvironment{CompactEnumerate}
  {\def\usecounter{\compactify\latexusecounter}
   \begin{enumerate}}
  {\end{enumerate}\let\usecounter=\latexusecounter}

\usepackage{ifthen}
\usepackage{fancyhdr}

\newfont{\dft}{phvb at 6pt}
\newfont{\mft}{phvro at 6pt}
\newfont{\df}{phvb at 9pt}
\newfont{\mf}{phvro at 9pt}
  {\begin{list}{$\bullet$}%
     {\setlength{\parsep}{0pt}%
      \setlength{\topsep}{0pt}%
      \setlength{\itemsep}{0pt}}}%
  {\end{list}}

\makeatletter

\global\def\section{\@startsection {section}{1}{\z@}%
                                   {-1.5ex \@plus -0.8ex \@minus -.1ex}%
                                   {0.6ex \@plus.2ex}
                                   {\normalfont\bfseries\scshape\fontsize{11}{13}\selectfont}}
\global\def\subsection{\@startsection{subsection}{2}{\z@}%
                                     {-1.25ex\@plus -0.8ex \@minus -.1ex}%
                                     {0.3ex \@plus .1ex}
                                     {\normalfont\bfseries\fontsize{10}{12}\selectfont}}
\global\def\subsubsection{\@startsection{subsubsection}{3}{\z@}%
                                     {-1ex\@plus -1ex \@minus -.1ex}%
                                     {0.1ex \@plus .1ex}
                                     {\normalfont\itshape\fontsize{10}{12}\selectfont}}

\usepackage{bibspacing}
\def\noeditingmarks{}
\newcommand{\textred}[1]{\textcolor{red}{#1}}
\ifx\noeditingmarks\undefined
   \newcommand{\pgwrapper}[2]{\textred{#1: #2}}
\else
   \newcommand{\pgwrapper}[2]{}
\fi

\usepackage{epsfig,comment}
\newcommand{\squishlist}{
   \begin{list}{$\bullet$}
    { \setlength{\itemsep}{0pt}      \setlength{\parsep}{3pt}
      \setlength{\topsep}{3pt}       \setlength{\partopsep}{0pt}
      \setlength{\leftmargin}{1.0em} \setlength{\labelwidth}{1em}
      \setlength{\labelsep}{0.5em} } }
\newcommand{\squishend}{
    \end{list}  }

\newcommand{\edit}[1]{{\textcolor{black}{#1}}}

\newcommand{\fillin}[1]{{\textcolor{black}{#1}}}

\date{}
\begin{document}

\twocolumn[\begin{@twocolumnfalse}

\begin{centering}

{\large \bf \name{}: Towards General-Purpose Acceleration of Retrospective Video Analytics\vspace{-0.1cm}\\}
{\vspace{0.3cm}
\begin{tabular}[t]{cc}
Neil Agarwal&Ravi Netravali\\
Princeton University&Princeton University
\end{tabular}\par}

\end{centering}

\vspace{\baselineskip}

\end{@twocolumnfalse}]

\interfootnotelinepenalty 100000
\widowpenalty 100000
\clubpenalty 100000
\newfont{\tf}{phvro at 9.5pt}
\newfont{\tft}{phvro at 7.25pt}
\begin{sloppypar}
\begin{abstract}
Commercial retrospective video analytics platforms have increasingly adopted general interfaces to support the custom queries and convolutional neural networks (CNNs) that different applications require. However, existing optimizations were designed for settings where CNNs were platform- (not user-) determined, and fail to meet at least one of the following key platform goals when that condition is violated: reliable accuracy, low latency, and minimal wasted work.

We present \name{}, a system that simultaneously meets all three goals while supporting the generality that today's platforms seek. Prior to queries being issued, \name{} carefully employs traditional computer vision algorithms to generate indices that are imprecise, but are fundamentally comprehensive across different CNNs/queries. For each issued query, \name{} employs new techniques to quickly characterize the imprecision of its index, and sparingly run CNNs (and propagate the results to other frames) in a way that bounds accuracy drops. Our results highlight that \name{}'s improved generality comes at low cost, with speedups that match (and most often, exceed) prior, model-specific approaches.

\end{abstract}
\section{Introduction}
\label{s:intro}

Video cameras are prevalent in our society, with massive deployments across major cities and organizations~\cite{LondonCamera,paris-hospital, beijing-cameras,ChicagoCamera,britain,va_market_mordor}. These cameras continually collect video data that is queried \emph{retrospectively} to guide traffic/city planning, business or sports analytics, healthcare, crime investigation, and many other applications~\cite{are-we-ready-for-ai-powered-security-cameras,powering-the-edge-with-ai-in-an-iot-world,vision-zero,smart-mall,traff2,fortune_va_market_report, analyzing-social-distancing, traffic-analysis, retail_example,sports_va,sports}. Queries typically involve running convolutional neural network (CNN) models that locate and characterize particular objects in scenes~\cite{pedestrian-detection-iccv15,cnn-face-cvpr15,pyramid-network-cvpr17,facial-point-cvpr13,imagenet-classification-cacm17}. Applications tailor the architectures and weights of those CNNs to their unique requirements (e.g., accuracy, latency, and resource cost) and target tasks, e.g., via specialization to scenes or object types~\cite{gemel,cornell_bird_cams_lab, custom_models_techsee}, proprietary training datasets~\cite{toyota, betterview, clearview}.

To support these diverse applications, commercial video analytics platforms (e.g., Microsoft Rocket~\cite{rocket,msft-cv,azure-face}, Amazon Rekognition~\cite{amazon-rekognition}, Google AI~\cite{google-cloud-vision}, IBM Maximo~\cite{ibm-maximo}) have steadily transitioned away from exposing only predetermined video processing results, towards being platforms that allow users/applications to register custom, large-scale video analytics jobs without worrying about infrastructural details~\cite{visor,privid,gemel}. To register a query, users typically provide (1) a CNN model of arbitrary architecture and weights, (2) a target set of videos (e.g., feeds, time periods), and (3) an accuracy target indicating how closely the provided results must match those from running the CNN on every frame. Higher accuracy targets typically warrant more inference (and thus, slower responses and higher costs).

From a platform perspective, there exist three main goals for each registered query. First and foremost, provided results should \emph{reliably} meet the specified accuracy target (usually above 80\%~\cite{gemel,reducto,focus-osdi18,blazeit-arxiv19}). Subject to that constraint, the platform should aim to consume as few computational resources as possible (i.e., minimize unnecessary work) and deliver responses as quickly as possible. The main difficulty in achieving these goals stems from the potentially massive number of video frames to consider, and the high compute costs associated with running a CNN on each one. For example, recent object detectors would require 500 GPU-hours to process a week of 30-fps video from just one camera~\cite{maskrcnn,speedacc}.

Unfortunately, despite significant effort in optimizing retrospective video analytics~\cite{focus-osdi18,tasti,blazeit,noscope-vldb17,visual-analytics-database-icde19,videozilla,otif}, no existing solution is able to simultaneously meet the above goals for the general interfaces that commercial platforms now offer. Most notably, recent optimizations perform ahead-of-time processing of video data to build indices that can accelerate downstream queries~\cite{focus-osdi18,tasti,otif}. However, these optimizations were designed for settings where models were known \emph{a priori} (i.e., not provided by users), and thus deeply integrate knowledge of the specific CNN into their ahead of time processing. Porting these approaches to today's bring-your-own-model platforms fundamentally results in unacceptable accuracy violations and resource overheads. The underlying reason is that models with even minor discrepancies (in architecture or weights) can deliver wildly different results for the same tasks and frames. Consequently, using different models for ahead-of-time processing and user queries can yield accuracy drops of up to 94\% (\S\ref{ss:limitations}). Building an index for all potential models is unrealistic given the massive space of CNNs~\cite{updating_nns, dataops, liu2019deep, zou2019object}, and the inherent risk of wasted resources since queries may never be issued~\cite{focus-osdi18,vstore-eurosys19}. 

In this paper, we ask \textit{``can retrospective video analytics platforms operate more like general-purpose accelerators to achieve their goals for the heterogeneous queries+models provided by users?''} We argue that they can, but doing so requires an end-to-end rethink of the way queries are executed, from the ahead-of-time processing used to develop indices, to the execution that occurs only once a user provides a model and accuracy target. We examine the challenges associated with each phase, and present \textbf{\name{}}, a complete video analytics platform that addresses those challenges.

\para{Ahead-of-time processing (indexing).} To support our goals, an index must meet the following criteria: (1) comprehensive with respect to data of interest for different models/queries -- any information loss would result in unpredictable accuracy drops, (2) links information across frames so CNN inference results -- the most expensive part of query execution~\cite{focus-osdi18,noscope-vldb17} -- can be propagated from one frame to another at low cost, and (3) cheap to construct since queries may never come in.

We show with \name{} that, if they are carefully applied in a \emph{conservative} manner, traditional computer vision (CV) algorithms~\cite{varcheie2010multiscale,urban_tracker_one, background-subtraction-cvpr2011,feature-tracks-iros16} can be repurposed to generate such an index per video.
Along these lines, \name{}'s ahead-of-time processing extracts a comprehensive set of \emph{potential} objects (or blobs) in each frame as areas of motion relative to the background scene. Trajectories linking blobs across frames are then computed by tracking low-level, model-agnostic video features, e.g., SIFT keypoints~\cite{sift}. Crucially, \name{}'s trajectories are computed once per video (not per video/model/query tuple) using cheap CV tasks that require only CPUs, and are generated 58\% faster than prior model-specific indices constructed using compressed CNNs and GPUs (\S\ref{ss:comparisons}).

\para{Query execution.} Once a user registers a query and CNN, the main question is how to use the comprehensive index to quickly generate results that meet the accuracy target, i.e., running inference on as few frames as possible, and aggressively propagating results along \name{}'s trajectories. The challenge is that \name{}'s index is extremely coarse and imprecise relative to CNN results. For instance, blob bounding boxes may be far larger than those generated by CNNs, and may include multiple objects that move in tandem. Worse, the imprecision of \name{}'s index varies with respect to different models and queries; prior systems avoid this issue by using indices that directly approximate specific models.

To handle this, \name{} introduces a new execution paradigm that first selects frames for CNN inference in a manner that sufficiently bounds the potential propagation error from index imprecision and unavoidable inconsistencies in CNN results~\cite{cnninconsistent}. The core idea is that such errors are largely determined by model-agnostic features about the video (e.g., scene dynamics), and can be discerned via inference on only a small set of representative frames. CNN results are then propagated using a custom set of accuracy-aware techniques that are specific to each query type (e.g., detection, classification) and robustly handle (and dynamically correct) imprecisions in \name{}'s trajectories. 

\para{Results.} We evaluated \name{} using 96 hours of video from 8 diverse scenes, a variety of CNNs, accuracy targets, and objects of interest, and three widely-used query types: binary classification, counting, and detection. Across these scenarios, \name{} consistently meets accuracy targets while running CNNs on only \fillin{3-54}\% of frames. Perhaps more surprisingly given its focus on generality and model-agnostic indices, \name{} outperforms existing systems that (1) rely solely on optimizations at query execution time (NoScope~\cite{noscope-vldb17}) by \fillin{19-97\%}, and (2) use model-specific indices (Focus~\cite{focus-osdi18} \edit{running with knowledge of the exact CNN}) by \fillin{-5-58\%}.

Taken together, our results affirmatively answer the question above, showing that \name{} can support the general interfaces and diverse user models that commercial platforms face, while delivering reliable accuracy and comparable (typically larger) speedups than prior, model-specific optimizations. We will open source \name{} post publication.

\section{Background and Motivation}
\label{s:motivation}

We begin with an overview of existing optimizations for retrospective video analytics (\S\ref{ss:existing}), and then present measurements highlighting their inability to generalize to the different models and queries that users register (\S\ref{ss:limitations}). We focus on the broad class of queries that use CNNs to characterize, label, and locate different types of objects in frames; \S\ref{s:overview} details the queries that \name{} supports. Such queries subsume those reported by both recent academic literature~\cite{blazeit-arxiv19,noscope-vldb17,privid,reducto,dds} and industrial organizations that run video analytics platforms~\cite{videostorm-nsdi17,focus-osdi18,gemel,chameleon-sigcomm18,ekya,optasia-socc16}. More concretely, we consider the following query types (and accuracy metrics):
\squishlist
\item \textbf{binary classification}: return a binary decision as to whether a specific object or type of object appears in each frame. Accuracy is measured as the fraction of frames tagged with the correct binary value. 

\item \textbf{counting}: return the number of objects of a given type that appear in each frame. Per-frame accuracy is set to the percent difference between the returned and correct counts.

\item \textbf{bounding box detection}: return the coordinates for the bounding boxes that encapsulate each instance of a specific object or object type. Per-frame accuracy is measured as the mAP score~\cite{Everingham:2010:PVO:1747084.1747104}, which considers the overlap (IOU) of each returned bounding box with the correct one.
\squishend

\begin{figure*}[t!]
\centering
\begin{subfigure}[t]{0.33\textwidth}
  \centering
  \includegraphics[width=1.01\textwidth]{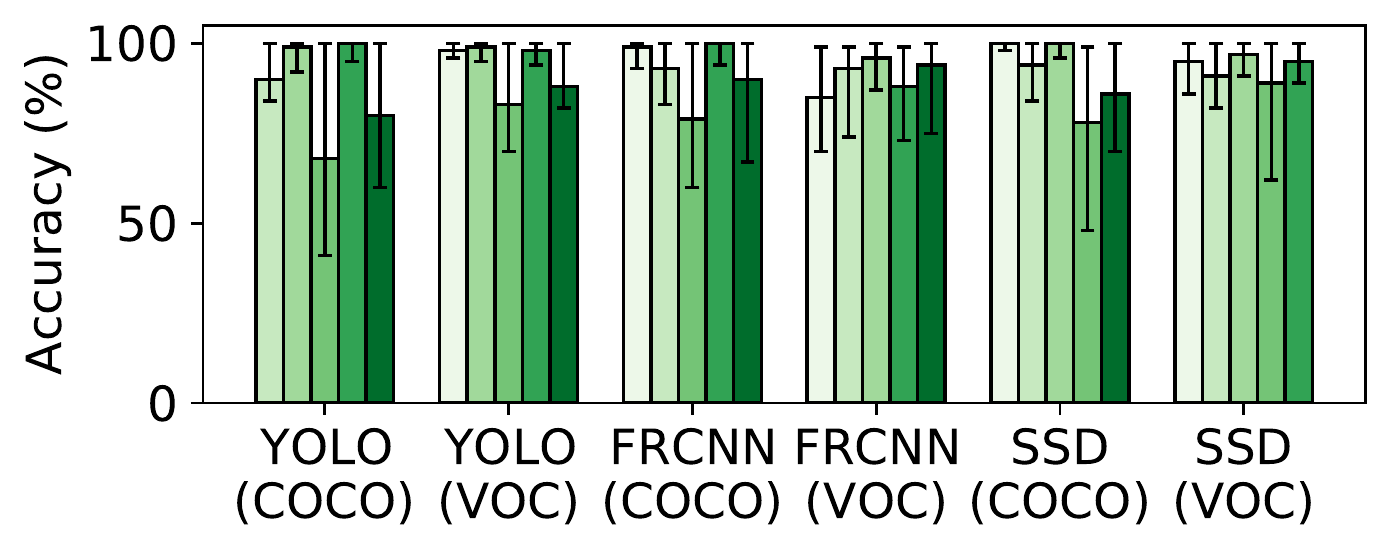}
 \vspace{-9pt}
  \tightcaption{Binary classification.}
\end{subfigure}
\begin{subfigure}[t]{0.33\textwidth}
  \centering
  \includegraphics[width=1.01\textwidth]{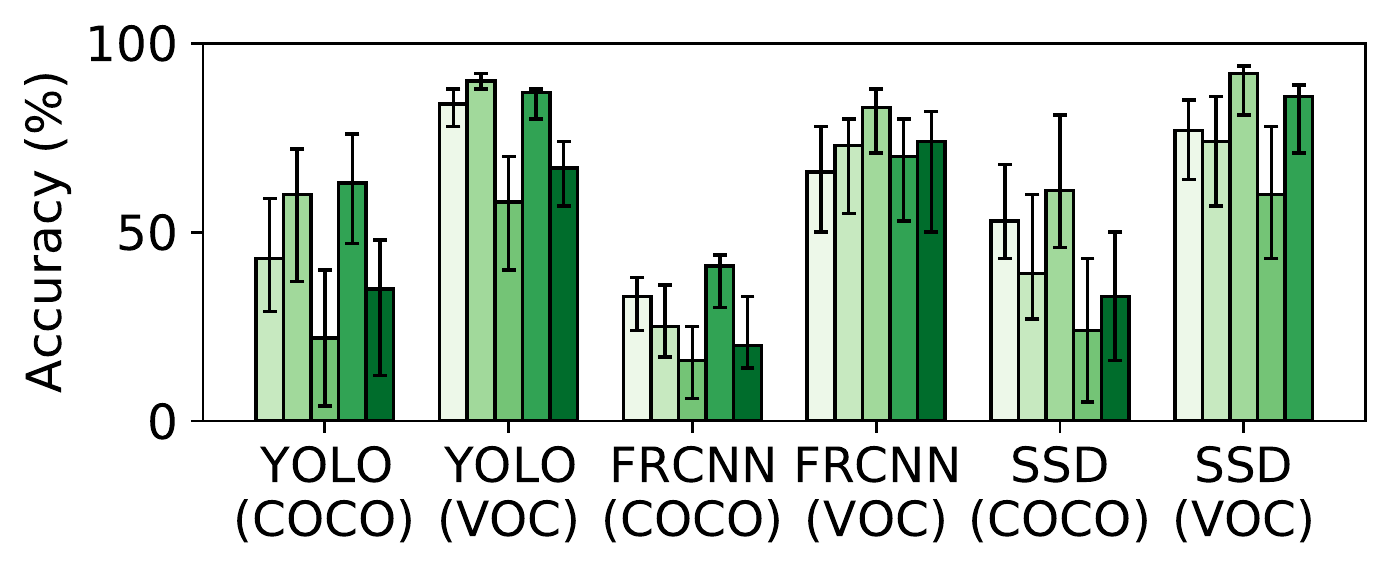}
  \vspace{-9pt}
  \tightcaption{Counting.}
\end{subfigure}
\begin{subfigure}[t]{0.33\textwidth}
  \centering
  \includegraphics[width=1.01\textwidth]{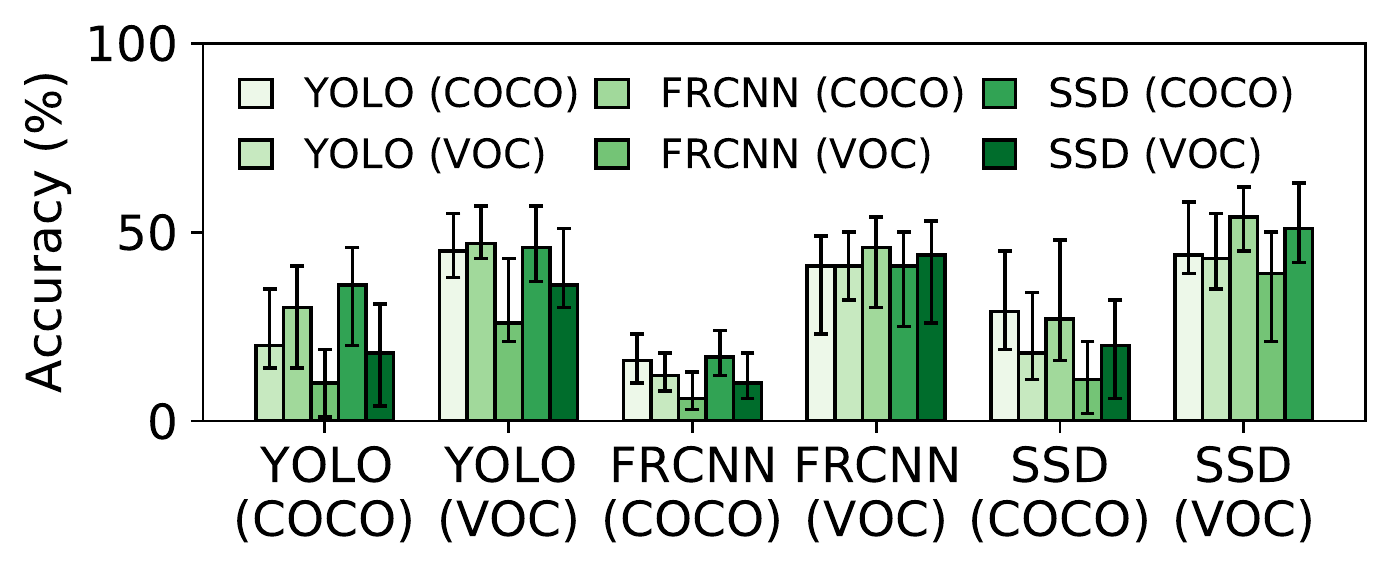}
  \vspace{-9pt}
  \tightcaption{Bounding box detection.}
\end{subfigure}
\vspace{-1pt}
    \tightcaption{Query accuracies when \emph{different} CNNs are used for preprocessing (bar types) and query execution (X axes). Bars show results for the median video, and error bars span the 25-75th percentiles. Models are listed as `architecture (training dataset)'.}
\label{fig:summary}
\end{figure*}

\subsection{Existing Acceleration Approaches}
\label{ss:existing}

\para{Query-time strategies.} Systems such as NoScope~\cite{noscope-vldb17} and Tahoma~\cite{visual-analytics-database-icde19} only operate once a user issues a query. To accelerate response generation, they first train cascades of cheaper binary classification CNNs that are specialized to the user-provided CNN, object of interest, and target video. The specific cascade to use is selected with the goal of meeting the accuracy target while minimizing computation and data loading costs. If confidence is lacking with regards to meeting the accuracy target, the user's CNN is incrementally run on frames until sufficient confidence is achieved.

\para{Ahead-of-time (preprocessing) strategies.}
Other systems provide speedups by performing some computation ahead of time, i.e., before a query is issued; for ease of disposition, we refer to such computations as \emph{preprocessing} in the rest of the paper. For example, Focus~\cite{focus-osdi18} speeds up binary classification queries by building an approximate, high-recall index of object occurrences using a specialized and compressed CNN that roughly matches the full CNN on the target video. Objects are then clustered based on the features extracted by the compressed model such that, during query execution, the full CNN only runs on the centroid of each cluster, with labels being propagated to all other objects in the same cluster.

BlazeIt~\cite{blazeit} and TASTI~\cite{tasti} accelerate aggregate versions of certain query types, e.g., total counts across all frames. Preprocessing for both systems involves generating sampled results using the full CNN. TASTI uses the sampled results to train a cheap embedding CNN that runs on all frames and clusters those that are similar from the model's perspective. During query execution, the full CNN is run only on select frames in each cluster, with the results propagated to the rest. In contrast, BlazeIt uses the sampled results to train specialized CNNs that act as control variates for the remaining frames: the specialized CNNs run on all frames, and the results are correlated with sampled results from  the full CNN to provide guarantees in statistical confidence. OTIF~\cite{otif} follows a similar strategy, but uses proxy models (trained using the sampled results) to extract tracks about model-specific objects that are later used to accelerate tracking queries.

Videozilla~\cite{videozilla} aims to extend such indexing optimizations across multiple video streams. More specifically, it identifies and exploits semantic similarities across streams that are based on the features extracted by the full CNN.

\subsection{The Problem: Model-Specific Preprocessing}
\label{ss:limitations}

As confirmed by prior work~\cite{focus-osdi18, tasti, blazeit} and our results in \S\ref{ss:comparisons}, preprocessing (intuitively) reduces the amount of computation required during query execution, and is crucial to enabling fast responses. However, \emph{all} existing solutions suffer from the same fundamental issue: they deeply integrate a specific CNN into their preprocessing computations (e.g., to generate sampled results for training the compressed models used to build indices or group similarly-perceived frames), and assume that all future queries will use that same exact CNN. While such an approach was compatible with prior platforms that exposed only predetermined results from platform-selected CNN(s), it is no longer feasible with the bring-your-own-model interfaces that are now commonplace on commercial platforms (\S\ref{s:intro}). To make matters worse, consider that queries can be made at any point in the future and the space of potential CNNs is immense and rapidly evolving~\cite{updating_nns, dataops, liu2019deep, zou2019object}, with variations in architecture (\eg \# of layers) or weights (\eg different training datasets).

To quantify the issues when this assumption is violated, we ran experiments asking: how would accuracy be affected if the CNN provided by users during query execution (i.e., \emph{query CNN}) was different than the CNN used during preprocessing (i.e., \emph{preprocessing CNN})? We consider the three query types above, videos and objects described in \S\ref{ss:methodology}, and a wide range of CNNs: Faster RCNN, YOLOv3, and SSD, each trained on two datasets (COCO and VOC Pascal).

For each possible pair of preprocessing and query CNNs, we ran both CNNs on the video to obtain a list of object bounding boxes per frame. In line with Focus' observation that classification results from two CNNs may not identically match but should intersect for the top-$k$ results~\cite{focus-osdi18}, we ignore the classifications from each CNN. Instead, we consider all bounding boxes from the preprocessing CNN that have an IOU of $\geq0.5$ with some box generated by the query CNN; results were largely unchanged for other IOU thresholds.  This presents the \emph{best scenario} (accuracy-wise) for existing preprocessing strategies.
Finally, we compute query results separately using only the remaining preprocessing CNN's boxes or all of the query CNN's boxes, and compare them.

\begin{figure}[t]
    \centering
    \includegraphics[width=0.8\columnwidth]{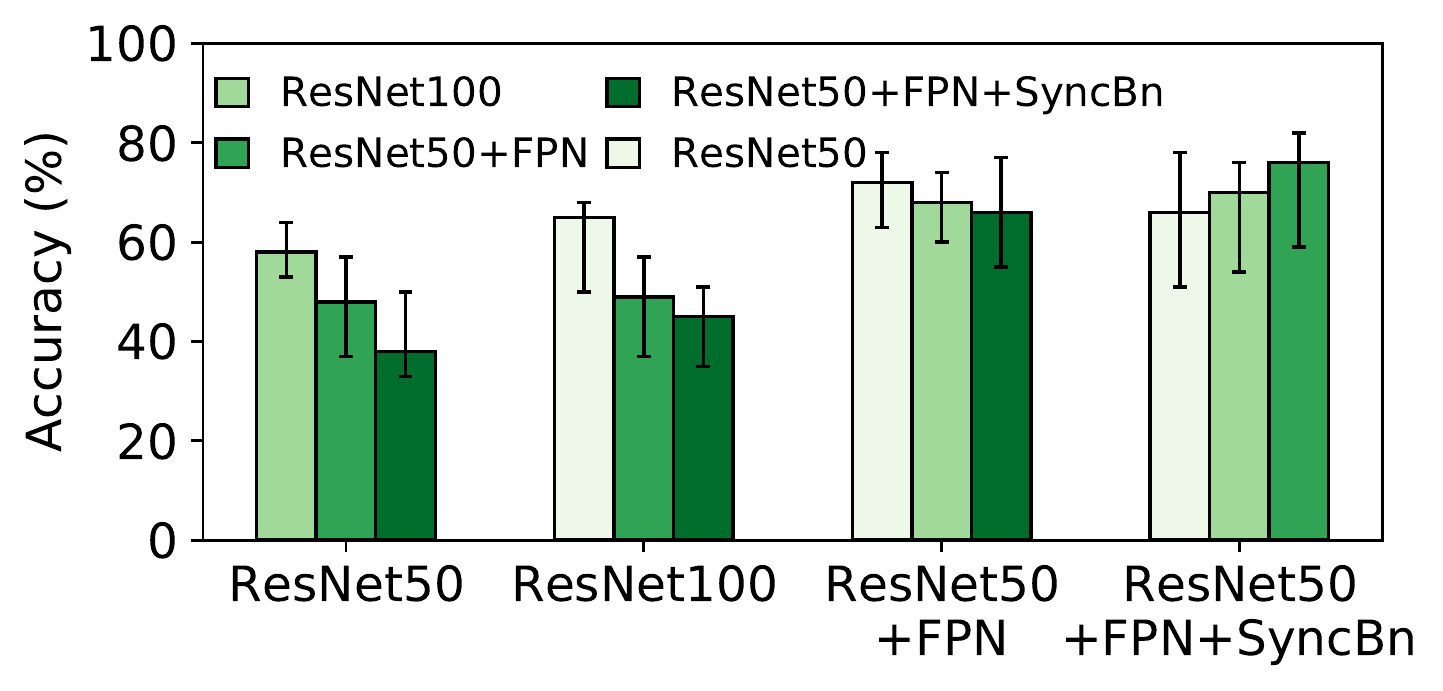}
    \tightcaption{Accuracies when CNNs for preprocessing (bar types) and query execution (X axis) are FasterRCNN+COCO with different ResNet backbones. Results are for counting queries; bars list medians with error bars for 25-75th percentiles.}
    \label{fig:faster_rcnn_variants}
\end{figure}

Figure~\ref{fig:summary} shows that discrepancies between preprocessing and query CNNs can lead to significant accuracy degradations, with the errors growing as query precision increases. For example, median degradations were \fillin{0-32}\% for binary classifications, but jump to \fillin{8-84}\% and \fillin{46-94}\% for counting and detections. Note that degradations for binary classification and counting
are by definition due to the preprocessing CNN entirely missing objects relative to the query CNN. Parsed differently, median degradations across query types were \fillin{0-84}\%, \fillin{2-94}\%, and \fillin{1-90}\% when the preprocessing and query CNNs diverged in terms of only architecture, only weights, or both. Figure~\ref{fig:faster_rcnn_variants} shows that these degradations persist even for variants in the same family of CNNs.

\para{Takeaway.} Ultimately, when run on the general interfaces of today's commercial video analytics platforms where users can provide CNNs, all existing optimizations would sacrifice at least one key platform goal:
\squishlist
\item \emph{reliable accuracy:} running preprocessing optimizations as is (using platform-determined CNNs) would yield unpredictable and substantial (up to 94\%) accuracy hits; 
\item \emph{minimal wasted work:} performing preprocessing for all potential user CNNs is not only unrealistic given the sheer number of possibilities, but would also result in substantial wasted work since queries may never be issued;
\item \emph{low-latency responses:} optimizing only once a query is issued will yield higher than necessary response times.
\squishend

\section{Overview of \name{}}
\label{s:overview}

\begin{figure*}[t]
    \centering
    \includegraphics[width=0.82\textwidth]{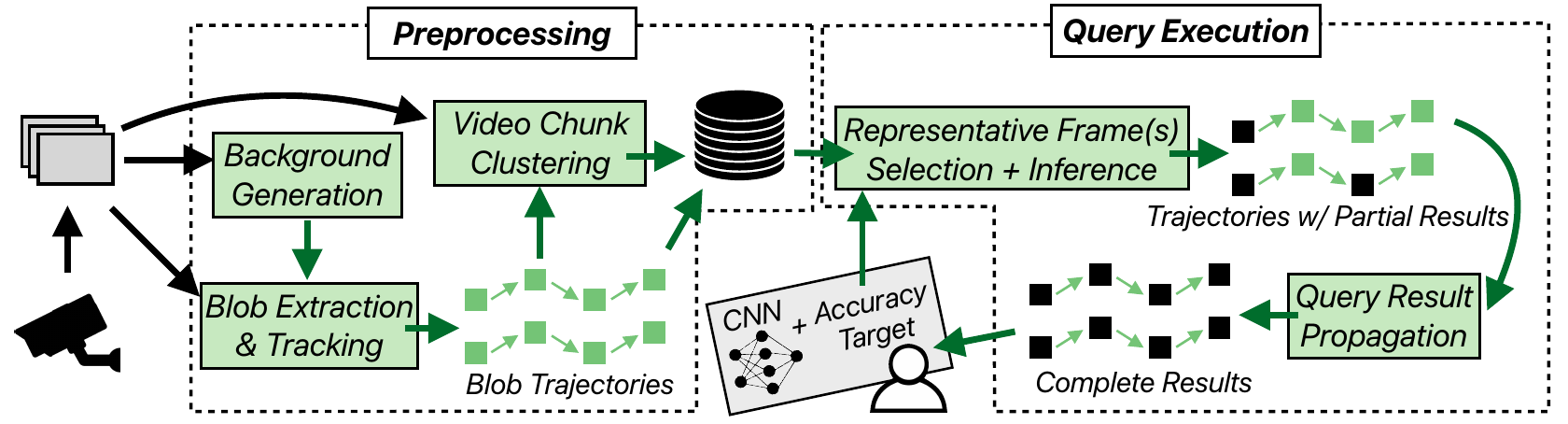}

    \vspace{0pt}
    \tightcaption{Overview of \name{}.}
    \label{fig:overview}
\end{figure*}

This section overviews the end-to-end workflow (Figure~\ref{fig:overview}) that \name{} uses to simultaneously meet all three platform goals for general, user-provided CNNs. We detail its preprocessing and query execution phases in \Sec{ingest}~and~\Sec{query}.

\para{Preprocessing.}
The main goal of \name{}'s preprocessing phase is to perform cheap computations over a video dataset such that the outputs (an index) can accelerate query execution for diverse user CNNs, without sacrificing accuracy. Crucially, to avoid the pitfalls of prior work (\S\ref{ss:limitations}), \name{}'s preprocessing does not incorporate \emph{any} knowledge of the specific CNN(s) that will be used during query execution. Instead, our insight is that traditional computer vision (CV) algorithms~\cite{stauffer99, varcheie2010multiscale,urban_tracker_one, background-subtraction-cvpr2011,feature-tracks-iros16} are well-suited for such preprocessing, as they extract information purely about video data, rather than how a specific model or query would parse that data. Using generic CV algorithms enables \name{} to generate a single index per video, rather than per video/query/model tuple. Further, those CV algorithms are computationally cheaper than (even compressed) CNNs, and rely on CPUs (not GPUs), keeping monetary costs low (\S\ref{ss:comparisons}). Both aspects drastically reduce the potential for wasted work.

However, in contrast to their intended use cases, for our purposes, CV algorithms must be \emph{conservatively} tuned to ensure that accuracy during query execution is not sacrificed. In other words, \name{}'s index must \emph{comprehensively} include all information that may influence or be incorporated in a query result (across CNNs), regardless of how coarse or imprecise that information is. Whereas coarse or imprecise results can be corrected or filtered out during query execution, missing information would result in unpredictable accuracy drops.

\edit{Accordingly, \name{} carefully uses a combination of motion extraction and low-level feature tracking techniques to identify all potential objects as areas of motion (or \emph{blobs}) relative to a background estimate, and record their \emph{trajectories} across frames by tracking each blob's defining pixels (or keypoints). For the former task, only high-confidence pixels are marked as being part of the background, ensuring that even minor motion is treated as a potential object; note that static objects are definitively discovered during query execution via CNN sampling on the frames across which the objects are static. For the latter task, any uncertainty in trajectory continuity (e.g., tracking ambiguities) is handled by simply starting a new trajectory; this ensures that results are not mistakenly propagated across different objects during query execution, albeit at the cost of additional inference.  Overall, we did not observe \emph{any} missed moving objects in \name{}'s indices across our broad evaluation scenarios (\S\ref{ss:methodology}).}

Trajectories are a fundamental shift from the clustering strategies that prior systems use to group frames or objects based on how they are perceived by a specific CNN (\S\ref{ss:existing}). In contrast, trajectories are computed in a model-agnostic manner, but still provide a mechanism through which to propagate CNN results across frames during query execution -- the primary source of speedups. Such generality does, however, come at a cost. Whereas prior efforts cluster on frames or object classes, \name{}'s trajectories group frames on a per-object basis. This discrepancy lets \name{} defer the determination of how a user's CNN perceives each object to query execution, but it limits the amount of potential result propagation, i.e., \name{} propagates the result for an object across the frames in which it appears, while prior approaches can propagate results across appearances of different objects. Note that this discrepancy does not apply to detection queries that require precise object locations (not just labels) to be propagated.

A natural question is: why not cluster objects on the features extracted by traditional CV algorithms to enable more result propagation? The issue is that, if performed without knowledge of the user-provided CNN, such clustering could lead to unsafe result propagation. More specifically, objects that are similar on some set of features but are perceived differently by the user's CNN could end up in the same cluster.

\para{Query Execution.}
Once a user registers a query (providing a CNN, accuracy target, and video to consider), \name{}'s goal is to generate a full set of per-frame results as quickly as possible, while reliably meeting the target accuracy. This translates to using the index from preprocessing (i.e., blobs and trajectories) to run the CNN on a small sample of frames, and efficiently propagate those results to the remaining frames.

The main challenge is that, owing to their general-purpose nature (relative to different models/queries) and closeness to noisy image signals, the CV algorithms used during preprocessing typically produce results that fail to precisely align with those from a user-provided CNN~\cite{reducto,privid}. Consequently, in being comprehensive,
\name{}'s index is coarse and imprecise relative to the target results from a user's CNN, e.g., with misaligned bounding boxes or extraneous objects that are not of interest to the query. Worse, the degree of imprecision is specific to the user CNN, and can lead to cascading errors (and accuracy drops) as results are propagated along \name{}'s trajectories. All prior efforts avoid these issues by tuning indices to specific CNNs at the cost of generality. 

To bound accuracy degradations (and reliably meet the specified target) while avoiding substantial inference, \name{} introduces a new query execution approach with two main components. First, to quickly and judiciously select the frames to run CNN inference on, our key observation is that errors from index imprecision and result propagation are largely dictated by model-agnostic features about the video, e.g., scene dynamics, trajectory lengths. Accordingly, \name{} clusters chunks of video in the dataset based on those features, and runs the user's CNN only on cluster centroids to determine the best frame selection strategy per cluster for the query at hand, i.e., the lowest frequency of CNN inference that meets the user-specified target accuracy. We note that, since clustering is based on model-agnostic features, it can be performed during preprocessing; CNN inference on centroids, however, only occurs once a user registers a query.

Second, to further limit inference overheads, \name{} introduces a new set of result propagation techniques that are specific to each query type and bolster propagation distances in spite of imprecisions in the index.  For instance, for bounding box detections, \name{} leverages our empirical observation that the relative position between an object’s keypoints (from preprocessing) and its bounding box edges remain stable over time. Building on this, \name{} propagates an object's CNN-produced bounding box to subsequent frames in its trajectory by efficiently searching for the coordinates that maximally preserve these spatial relationships.

\para{Query model and assumptions.} \name{} currently supports the large body of object-centric queries whose results are reported at the granularity of individual objects (e.g., labeling or locating them) and whose CNNs are run on a per-frame basis. Thus, currently handled queries include classifications, counting, and detections, as well as queries that build atop those primitives such as tracking and activity recognition.
Such queries dominate the workloads reported by commercial platforms~\cite{videostorm-nsdi17, focus-osdi18, gemel, chameleon-sigcomm18, ekya}, and subsume those supported by prior work (\S\ref{ss:existing}). We note that \name{}'s approach is general enough to also accelerate less common, finer-grained queries, \eg semantic segmentation~\cite{semantic_segmentation}. For such queries, the keypoints (and their matches across frames) recorded in \name{}'s index can be used to propagate groups of pixel labels; we leave implementing this to future work.

\name{} does not make any assumptions about or require any knowledge of the object type(s) that a query targets. Instead, as described above, \name{} relies on generic background estimation and motion extraction to identify potential objects in the scene. The intuition is that a moving object of any kind will involve (spatially correlated) moving pixels that can be identified purely based on the scene. \name{} leaves it to the user's CNN to determine whether those potential objects are of interest during query execution. We stress-test \name{}'s robustness to different object types in \S\ref{ss:deepdive}.

\name{}'s preprocessing currently operates on videos from static cameras. We note, however, that the CV community has actively been extending the core techniques that \name{} builds atop to deliver improved robustness in the face of moving cameras~\cite{moving_cam1,moving_cam2,moving_cam3,moving_cam4}. We leave an exploration of integrating these efforts into \name{} to future work.

\section{\name{}'s Preprocessing}
\label{s:ingest}

\begin{figure*}[t]
    \centering
    \includegraphics[width=0.99\textwidth]{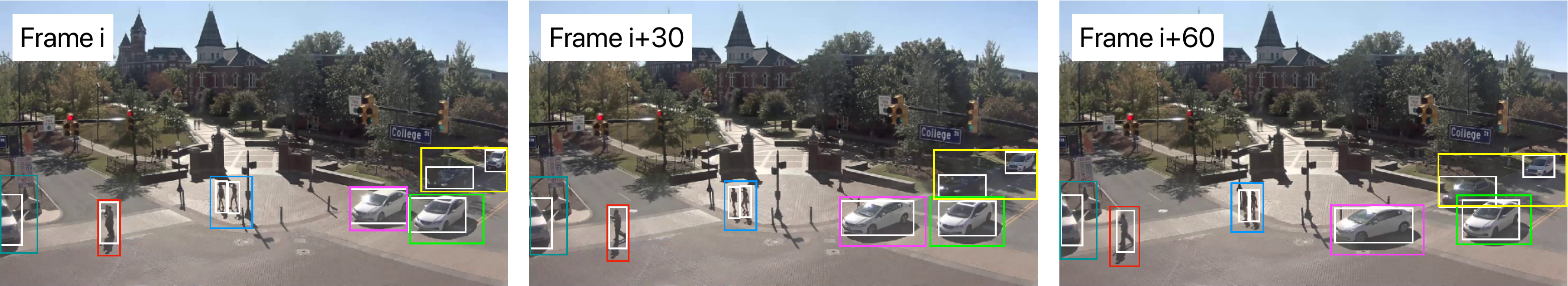}
    \vspace{2pt}
    \tightcaption{Example screenshots from the Auburn video (Table~\ref{t:dataset}). CNN (YOLOv3+COCO) detections are shown in white, while each of \name{}'s trajectories (and their constituent blobs) is shown in a different color.}
    \label{fig:example_screenshots}
\end{figure*}

\name{}'s target output from preprocessing is a set of blobs and their trajectories. To efficiently extract this information and enable parallel processing over the dataset, \name{} operates independently on video chunks (i.e., groups of contiguous frames); the default chunk size is 1 min (profiled in \S\ref{ss:deepdive}), and trajectories are bound to individual chunks to eliminate any cross-chunk state sharing. The rest of this section describes the analysis that \name{} performs per chunk.

\para{Background estimation.}
Extracting blobs inherently requires a point of reference against which to discern areas of motion. Thus, \name{}'s first task is to generate an estimate of the background scene for the current chunk. However, existing background estimation approaches~\cite{laugraud2017labgen, barnich2010vibe} are ill-suited for \name{} as they are primarily concerned with generating a single, coherent background image despite scene dynamics (e.g., motion) that complicate perfect foreground-background separation. In contrast, \name{}'s focus is on navigating the following tradeoff between accuracy and efficiency, not coherence. On one hand, placing truly background pixels in the foreground will lead to spurious trajectories (and inefficiencies during query execution). On the other hand, incorrectly placing a temporarily static object in the background can result in accuracy degradations. Indeed, unlike entirely static objects that will surely be detected via CNN sampling and propagated to all frames in a chunk (during query execution), temporarily static objects may be missed and should only be propagated to select frames.

\name{} addresses the above tradeoff in a manner that favors accuracy. More specifically, \name{} only marks content as pertaining to the background scene when it has high confidence; all other content is conservatively marked as part of the foreground and is resolved during query execution. To realize this approach, \name{} eschews recent background estimation approaches in favor of a custom, lightweight strategy.

In its most basic form, background estimation involves recording the distribution of values assigned to each pixel (or region) across all frames in the chunk, and then marking the most frequently occurring value(s) (i.e., the peaks in the probability density function) as the background~\cite{stauffer99, varcheie2010multiscale}. This works well in scenarios where there is a clear peak in the distribution that accounts for most of the values, e.g., if objects do not pass through the pixel or do so with continuous motion, or if an object is entirely static and can thus be safely marked as the background. However, complications arise in settings with multiple peaks. For instance, consider a pixel with two peaks. Any combination of peaks could pertain to the background: a tree could sway back and forth (both), a single car could temporarily stop at a traffic light (one), or multiple cars could serially stop-and-go at the light (none).

To distinguish between these multi-modal cases and identify peaks that definitely pertain to the background for a chunk, \name{} extends (into the next chunk) the duration over which the distribution of pixel values is computed. The idea is that motion amongst background components should persist as we consider more video, while cases with temporarily static object should steadily transform into uni-modal distributions favoring either the background scene or the object (if it remains static). To distinguish between the object and background in the latter case, \name{} further extends the distribution of pixel values to incorporate video from the previous chunk. If the same peak continues to rise, it must pertain to the background since we know that the object was not static throughout the entire chunk. Otherwise, \name{} conservatively assigns an empty background for that pixel.

\para{Blob Extraction.}
Using the background estimate, \name{} takes a second pass through the chunk in order to extract areas of motion (blobs) on each frame. More specifically, \name{} segments each frame into a binary image whereby each pixel is annotated with a marker specifying whether it is in the foreground or background. Our implementation deems a pixel whose value falls within 5\% of its counterpart(s) in the background estimate as a background pixel, but we find our results to be largely insensitive to this parameter. Given the noise in low-level pixel values~\cite{reducto}, \name{} further refines the binary image using a series of morphological operations~\cite{morph}, e.g., to convert outliers in regions that are predominantly either background or foreground. Lastly, \name{} derives blobs by identifying components of connected foreground pixels~\cite{granabbdt}, and assigning a bounding box using the top left and bottom right coordinates of each component.

\para{Computing Trajectories.}
\name{}'s final preprocessing task is to convert the set of per-frame blobs into trajectories that track each blob across the video chunk. At first glance, it may appear that sophisticated multi-object trackers (\eg Kalman Filters)~\cite{bewley2016simple, wojke2017simple, fancy_tracking1, fancy_tracking2} could directly perform this task. However, most existing trackers rely on pristine object detections as input. Blobs do not meet this criteria, and instead are far coarser and imprecise (Figure~\ref{fig:example_screenshots}). At any time, a single blob may contain multiple objects, e.g., two people walking together. Blobs may split or merge as their constituent objects move and intersect. Lastly, the dimensions of a given object's blob bounding boxes can drastically fluctuate across frames based on interactions with the estimated background.

To handle these issues, we turn to tracking algorithms that incorporate low-level feature keypoints (SIFT~\cite{sift} in particular)~\cite{urban_tracker_one,urban_tracker_two}, or pixels of potential interest in an image, e.g., the corners that \emph{may} pertain to a car windshield. Associated with each keypoint is a descriptor that incorporates information about its surrounding region, and thus enables the keypoint (and its associated content) to be matched across images. \name{} conservatively applies this functionality to generate correspondences between blobs across frames. 

For each pair of consecutive frames, \name{} pairs the constituent keypoints of each blob. This may yield any form of an N $\rightarrow$ N correspondence depending on the underlying tracking event, e.g., blobs entering/leaving a scene, fusion or splitting of blobs. For instance, if the keypoints in a blob on frame $f_i$ match with keypoints in N different blobs on frame $f_{i+1}$, there is a 1 $\rightarrow$ N correspondence. To generate trajectories, \name{} makes a series of forwards and backwards scans through the chunk. For each correspondence that is not 1 $\rightarrow$ 1, \name{} propagates that information backwards to account for the observed merging or splitting. For example, for a 1 $\rightarrow$ N correspondence between frames $f_{i}$ and $f_{i+1}$, \name{} would split $f_{i}$'s blob into N components using the relative positions of the matched keypoints on $f_{i+1}$ as a guide.

\para{Index Storage.} Preprocessing outputs are stored in MongoDB~\cite{mongodb}; costs profiled in \S\ref{ss:deepdive}. Matched keypoints are stored with the corresponding frame IDs: row = [$<$((x,y)-coordinates, frame \#)$>$]. Blob coordinates (and their trajectory IDs) are stored per frame to facilitate the matching of CNN results and blobs on sampled frames during query execution (\S\ref{ss:propagate}): 
row = [$<$((x,y)-coordinates of top left corner, (x,y)-coordinates of bottom right corner, trajectory ID)$>$].
\section{Fast, Accurate Query Execution}
\label{s:query}

During query execution, \name{}'s sole goal is to judiciously use the user-provided CNN and the index from preprocessing to quickly generate a complete set of results that meet the specified accuracy target. Doing so involves answering two questions: (1) what sampled (or \emph{representative}) frames should the CNN be run on such that we can sufficiently adapt to the registered query (i.e., CNN, query type, and accuracy target) and bound errors from index imprecisions?, and (2) how can we use preprocessing outputs to accurately propagate sampled CNN results across frames for different query types? For ease of disposition, we describe (2) first, assuming CNN results on representative frames are already collected.

\subsection{Propagating CNN Results}
\label{ss:propagate}

Regardless of the query type, \name{}'s first task is to pair the CNN's bounding box detections on representative frames with the 
blobs on those same frames; this, in turn, associates detections with trajectories, and enables cross-frame result propagation. To do this, we pair each detection bounding box with the blob that exhibits the maximum, non-zero intersection. Trajectories that are not assigned to any detection are deemed spurious and are discarded. Further, detections with no matching blobs are marked as `entirely static objects' and are handled after all other result propagation (described below). Note that, with this approach and in spite of the trajectory corrections from \S\ref{s:ingest}, multiple detections could be associated to a single blob, i.e., when objects move together and never separate. Using these associations, \name{} propagates CNN results via techniques specific to the target query type.

\para{Binary classification and counting.}
To support both query types, each trajectory is labeled with an object count according to the number of detections associated with it on representative frames. If a trajectory passes through multiple representative frames, \name{} partitions the trajectory into segments, and assigns each segment a count based on the associations from the closest representative frame. Lastly, \name{} sums the counts across the trajectories that pass through each frame, and returns either the raw count (for counting), or a boolean indicating if count $>0$ (for binary classification).

\begin{figure}[t]
    \centering
    \includegraphics[width=0.9\columnwidth]{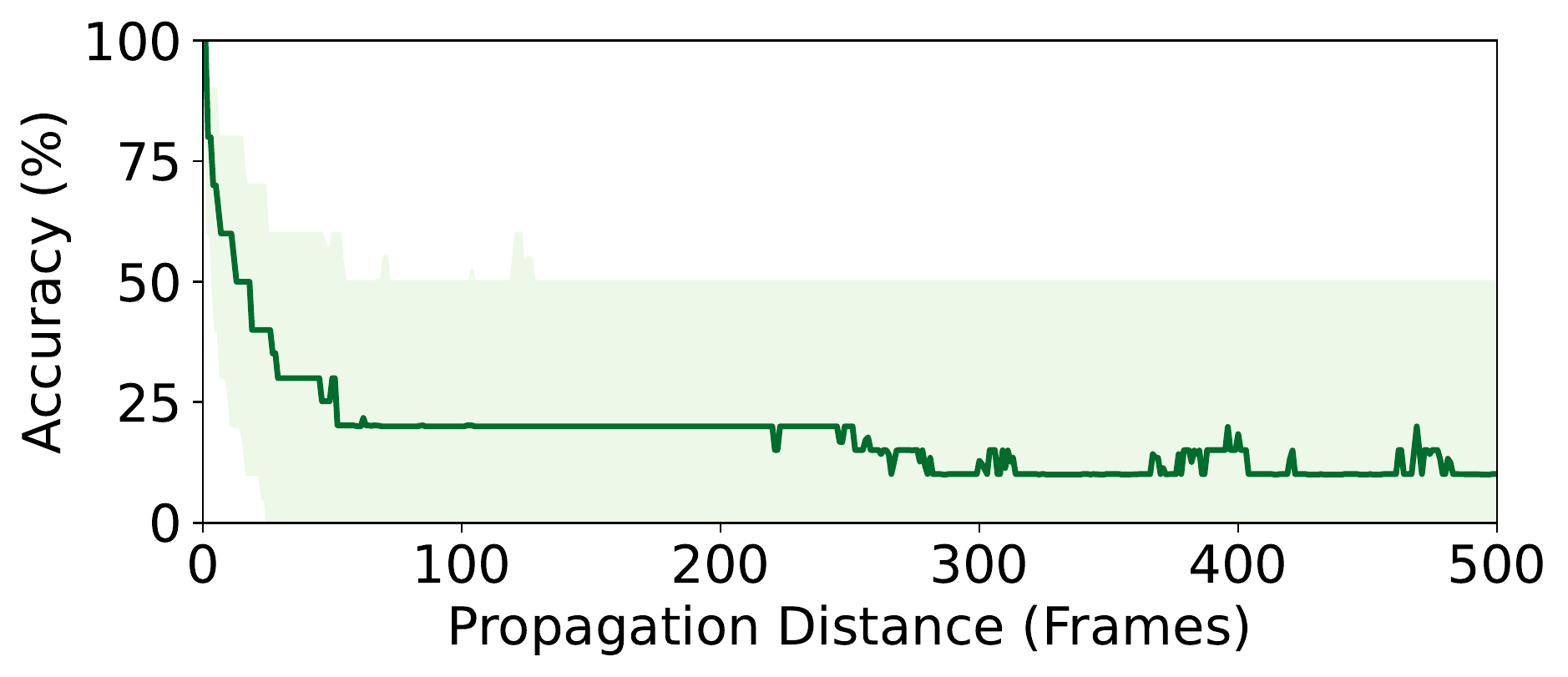}
    \tightcaption{Accuracy (mAP) degradations when CNN bounding boxes are propagated by computing the blob-to-detection coordinate transformation on a representative frame, and applying it to all other blobs in the trajectory. Line represents median detections, with ribbons spanning 25-75th percentiles.}

    \label{fig:coordinate_transforms_bad}
\end{figure}

\para{Bounding box detections.}
Whereas binary classification and count queries simply require propagating coarse information about object presence, bounding box queries require precise positional information to be shared across frames. However, as noted in \S\ref{s:ingest}, blobs and trajectories are inherently imprecise and fail to perfectly align with detections. A natural approach to addressing such discrepancies is to compute coordinate transformations between paired detections and blobs on representative frames, and apply those transformations to the remainder of each blob's trajectory; equivalently, one could compute transformations for a blob across its own trajectory, and apply them to add detections to non-representative frames. Unfortunately, Figure~\ref{fig:coordinate_transforms_bad} shows that detection accuracy rapidly degrades with this approach, e.g., median degradations are \fillin{30}\% when propagating a box over 30 frames. The reason is that blobs and their paired detections move/resize differently across frames, resulting in median errors of \fillin{84}\% between the Euclidean distances of blob-blob and detection-detection coordinate transformations.

\begin{figure}[t]
    \centering
    \includegraphics[width=.9\columnwidth]{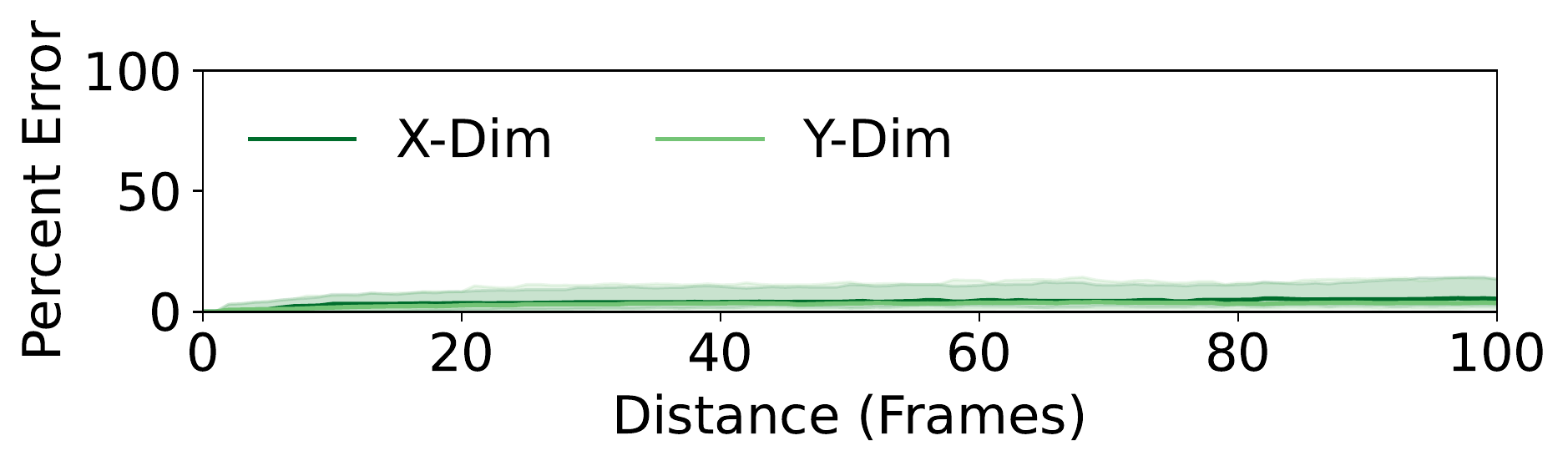}
    \tightcaption{Percent difference in anchor ratios for each object's keypoints across its trajectory. Line shows medians, with ribbons spanning 25-75th percentiles.}
    \vspace{2pt}
    \label{fig:keypoints_stable}
\end{figure}

To fill the void of stable propagation mechanisms, \name{} leverages our finding that the relative positions between an object's constituent keypoints (i.e., those extracted and tracked during preprocessing) and its detection bounding box edges remain largely unchanged over short durations; we refer to these relative positions as \emph{anchor ratios} since they `anchor' an object's content to a relative position within the bounding box. This stability is illustrated in Figure~\ref{fig:keypoints_stable}, and is intuitive: objects tend to remain rigid over short time scales, implying that the points they are composed of move in much the same way as the entire object does (including as the object scales in size). Building on this, \name{} propagates detections by using matching keypoints along the trajectories to which they have been associated, and efficiently solving an optimization problem in search of bounding box coordinates that maximally preserve the anchor ratios for each keypoint. 

More formally, for each detection on each representative frame, \name{} considers the set of keypoints $K$ that fall in the intersection with the associated blob. Each keypoint $k$ in $K$ has coordinates $(x_k,y_k)$. Further, let the coordinates of the detection bounding box be $(x_1, y_1, x_2, y_2)$, where $(x_1, y_1)$ and $(x_2, y_2)$ refer to the top left and bottom right corners. The anchor ratios $(ax_k, ay_k)$ for keypoint $k$ are computed as:
\vspace{-0.1cm}
\begin{equation}
    (ax_k, ay_k) = \Big(\frac{x_2-x_k}{x_2-x_1},\frac{y_2-y_k}{y_2-y_1}\Big)
\end{equation} 

For each subsequent non-representative frame (until the next representative frame) that includes the same trajectory, \name{} finds the set of keypoints that match with those in $K$; denote the set of matching keypoints as $K'$, where each $k'$ in $K'$ matches with keypoint $k$ in $K$. Finally, to place the bounding box on the subsequent frame, \name{} solves for the corresponding coordinates $(x_1, y_1, x_2, y_2)$ by minimizing the following function to maximally preserve anchor ratios:
\vspace{-0.2cm}
\begin{equation} \label{eq:op_fn}
   \mathlarger{\sum}^{K'}_{k'}{\Bigg[\Big(\frac{x_2-x_{k'}}{x_2-x_1}-ax_k\Big)^2 + \Big(\frac{y_2-y_{k'}}{y_2-y_1}-ay_k\Big)^2\Bigg]}
\end{equation}
\vspace{-0.2cm}

Note that this optimization (which takes 1 ms for the median detection) can be performed in parallel across frames and across detections on the same frame. Further, \name{} initializes each search with the coordinates of the corresponding detection box on the representative frame, thereby reducing the number of steps to reach a minima.

\para{Propagating entirely static objects.} Thus far, we have only discussed how to propagate detection bounding boxes that map to a blob/trajectory, i.e., moving objects. However, recall from \S\ref{s:ingest} that certain objects which are entirely static will be folded into the background scene. These objects are discovered by the CNN on representative frames, but they will not be paired with any blob. Instead, \name{} broadcasts these objects to nearby frames (until the next representative frame) in a query-specific manner: such objects add to the per-frame counts used for classification and count queries, and their boxes are statically added into frames for detection queries.

\subsection{Selecting Representative Frames}
\label{ss:repframes}

\begin{figure}[t]
    \centering
    \includegraphics[width=0.9\columnwidth]{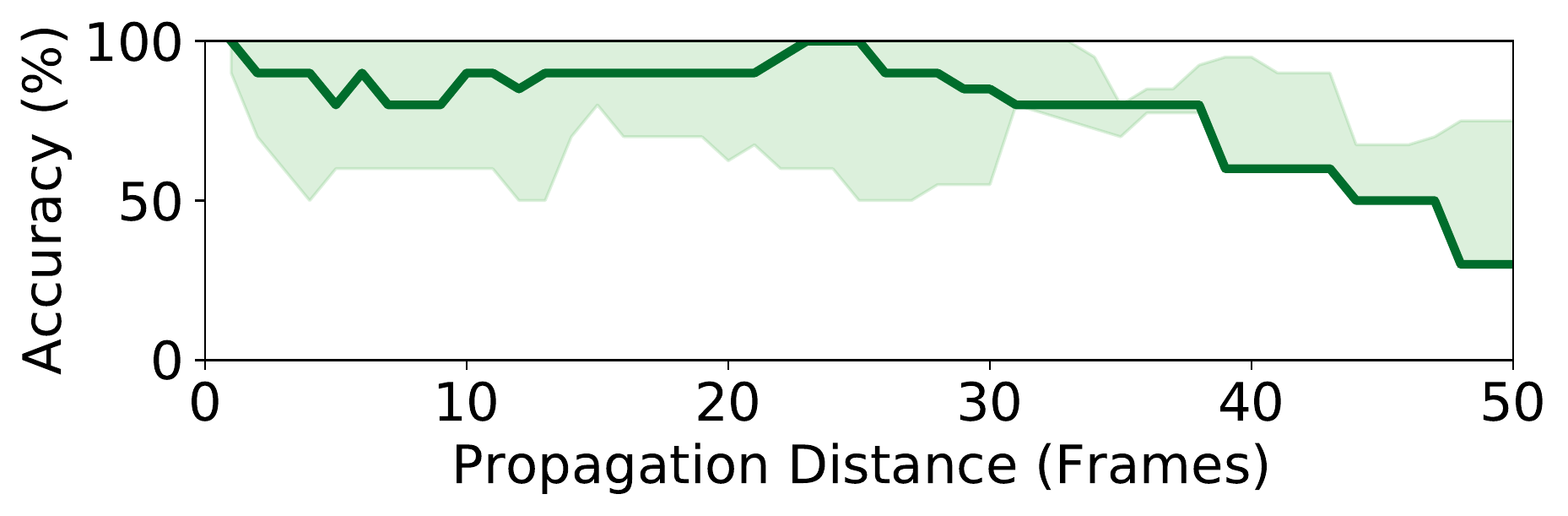}
    \vspace{1pt}
    \tightcaption{Accuracy (mAP) degradations grow as \name{} propagates detection bounding boxes over longer durations. Results consider all object trajectories in the median video. Line represents medians, with ribbons for 25-75th percentiles.}
    \label{fig:prop_error}
\end{figure}

To use the result propagation techniques from \S\ref{ss:propagate}, we must determine the set of sampled, representative frames to collect CNN results on. Because CNN execution is by far the largest contributor to query execution delays (\S\ref{ss:deepdive}), we aim to select the smallest set of representative frames such that \name{} can sufficiently discern the relationship between its index and the CNN results, and generate a complete set of accurate results.

A natural strategy for selecting representative frames is to pick the smallest set of frames such that every trajectory appears at least once. In theory, executing the CNN on this set of frames should be sufficient to generate a result (e.g., object label, bounding box) for each trajectory, and propagate that result to all of the trajectory's frames. However, this straightforward approach falls short for two reasons:
\begin{CompactEnumerate}

\item CNNs can be inconsistent and occasionally produce different results for the same object across frames, e.g., a car in frame $i$ may be ignored in frame $i+1$~\cite{cnninconsistent,kang2016object}. In line with prior analyses, we mostly observe this behavior for small or distant objects, e.g., YOLOv3 mAP scores are 18\% and 42\% for the small and large objects in the COCO dataset~\cite{redmon2018yolov3}. The consequence is that, if such an inconsistent result appears on a representative frame, \name{} would propagate it to all other frames in the trajectory, thereby spreading the error.

\item Even for consistent CNN results, propagation errors inherently grow with longer trajectories (i.e., as a given result is propagated to more frames). For instance, median accuracies are \fillin{90}\% and \fillin{30}\% when \name{} propagates bounding boxes over \fillin{10} and \fillin{50} frames (Figure~\ref{fig:prop_error}).

\end{CompactEnumerate}

These issues are more pronounced in busy/dynamic scenes with significant object occlusions/overlap~\cite{wang2018repulsion, hoiem2012diagnosing}. Moreover, the implication of both is that solely ensuring that the set of representative frames covers each trajectory is insufficient and can result in unacceptable accuracy degradations. To address this, \name{} introduces an additional constraint to the selection of representative frames: any blob in a trajectory must be within $max\_distance$ frames of a representative frame that contains the same trajectory. This, in turn, bounds both the duration over which inconsistent CNN results can be propagated, as well as the magnitude of propagation errors.

Tying back to our original goal, we seek the largest $max\_distance$ (and thus, fewest representative frames) that allows \name{} to meet the accuracy target. However, the appropriate $max\_distance$ depends on how the above issues manifest with the current query, CNN, and video. Digging deeper, we require an understanding of how \name{}'s propagation techniques (for the query type at hand) and the user's CNN interact with each frame and trajectory, i.e., how accurate are \name{}'s propagated results compared to the CNN's results. Though important for ensuring sufficient accuracy, collecting this data (particularly CNN results) for each frame during query execution would forego \name{}'s speedups. 

To achieve both accuracy and efficiency, \name{} clusters video chunks based on properties of the video and its index that characterize the aforementioned issues. The idea is that the chunks in each resulting cluster should exhibit similar interactions with the CNN and \name{}'s result propagation, and thus should require similar $max\_distance$ values. Accordingly, \name{} could determine the appropriate $max\_distance$ for all chunks in a cluster by running the CNN and result propagation only on the cluster's centroid chunk. 

To realize this approach, for each chunk, \name{} extracts distributions of the following features: object sizes (i.e., pixel area per blob), trajectory lengths (i.e., number of frames), and busyness (i.e., number of blobs per frame and trajectory intersections). These match our observations above: CNN inconsistencies are most abundant in frames with small objects, the potential for propagation errors is largest with long trajectories, and both issues are exacerbated in busy scenes.

\begin{figure}[t]
    \centering
    \includegraphics[width=0.9\columnwidth]{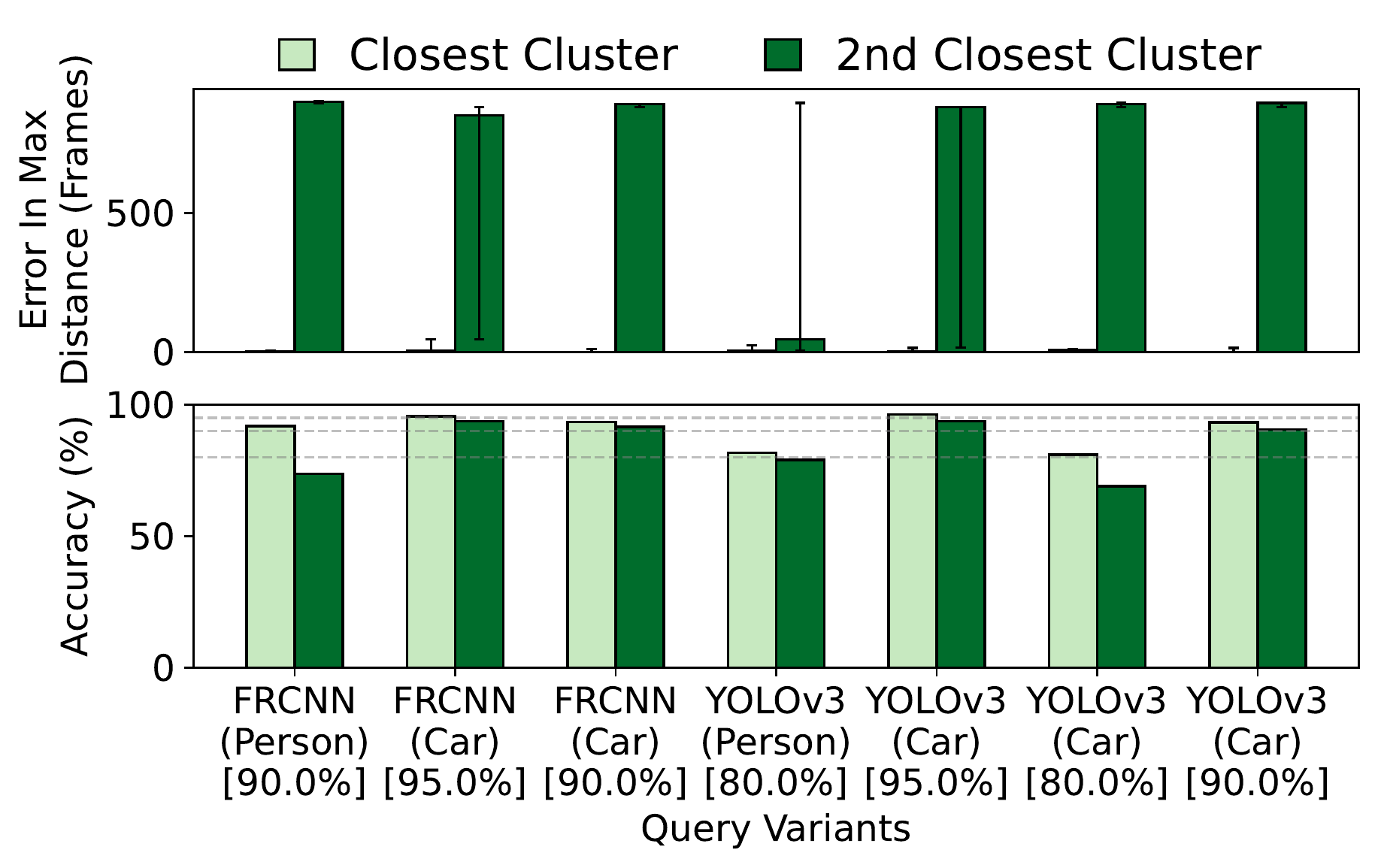}
    \vspace{1pt}
    \tightcaption{Effectiveness of \name{}'s clustering with different CNNs, (object types), and [accuracy targets]. Results are for the median video, and compare the ideal $max\_distance$ value for each chunk with those of the centroids in its cluster and the nearest neighboring cluster. The top graph measures the discrepancies in per-chunk $max\_distance$ (bars list medians, with error bars for 25-75th percentile); the bottom graph evaluates the corresponding hits on average accuracy (for detections).}
    \label{fig:clustering_graphs}
\end{figure}

With these features, \name{} clusters chunks using the K-means algorithm. We find that setting the number of target clusters to ensure that the centroids cover 2\% of video strikes the best balance between CNN overheads and robustness to diverse and rapidly-changing video chunks; we profile this parameter in \S\ref{ss:deepdive}. For each resulting cluster, \name{} runs the CNN on all frames in the centroid chunk. Using the collected results, \name{} runs its result propagation for a range of possible $max\_distance$ values, and computes an achieved accuracy for each one relative to the ground truth CNN results. More precisely, for each $max\_distance$, \name{} selects the set of representative frames by greedily adding frames until our criteria is met, i.e., all blobs are within $max\_distance$ of the closest representative frame containing the same trajectory. From there, \name{} selects the largest $max\_distance$ that meets the specified accuracy goal, and applies it to pick representative frames for all other chunks in the same cluster.

Figure~\ref{fig:clustering_graphs} highlights the effectiveness of \name{}'s clustering strategy in terms of (quickly) adapting to different query types, accuracy targets, objects of interest, and CNNs. As shown in Figure~\ref{fig:clustering_graphs}(top), the median discrepancy between each chunk's ideal $max\_distance$ value and that of the corresponding cluster centroid is only \fillin{0-8} frames; this jumps to \fillin{45-898} frames when comparing chunks with the centroid of the closest neighboring cluster. Figure~\ref{fig:clustering_graphs}(bottom) illustrates the importance of shrinking these discrepancies. More specifically, applying each centroid's ideal $max\_distance$ to all chunks in the corresponding cluster (i.e., \name{}'s approach) yields average accuracies that are consistently above the targets. The same is not true when using the ideal $max\_distance$ values from the nearest neighboring cluster.
\section{Evaluation}
\label{s:eval}

We evaluated \name{} on a wide range of queries, CNNs, accuracy targets, and videos. Our key findings are:

\squishlist

\item \name{} consistently meets accuracy targets while running the CNN on only \fillin{3-54\%} of frames, highlighting its comprehensive (model-agnostic) index and effective adaptation during query execution.

\item Despite its goal of generality, \name{}'s response times are \fillin{19-97}\% lower than NoScope's. Compared to Focus (which requires a priori knowledge of the CNN that will be used during query execution), \name{}'s response times are \fillin{33\%} and \fillin{52\%} lower on counting and detection queries, and only \fillin{5\%} higher on classifications.

\item \name{}'s preprocessing (and index construction) runs \fillin{58}\% faster than Focus', while also generalizing to different CNNs/queries and requiring only CPUs (not GPUs).

\item \name{}'s preprocessing and query execution tasks speed up nearly linearly with increasing compute resources.

\squishend

\begin{figure*}[t!]
    \centering
    \begin{subfigure}[t]{0.33\textwidth}
        \centering
        \includegraphics[width=\columnwidth]{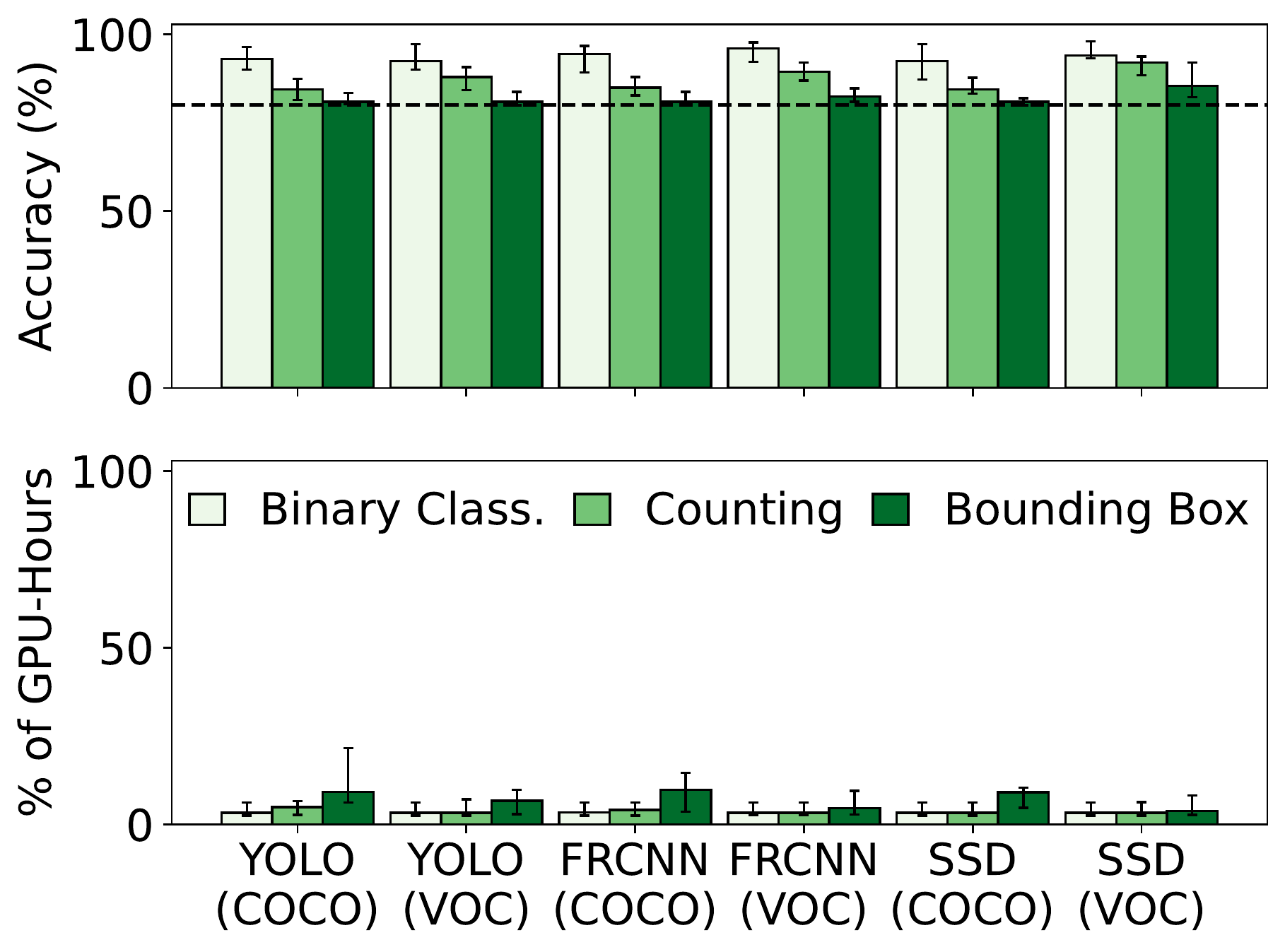}
        \label{fig: main_result_a}
        \vspace{-8pt}
        \tightcaption{80\% Accuracy Target}
    \end{subfigure}
    \begin{subfigure}[t]{0.33\textwidth}
        \centering
        \includegraphics[width=\columnwidth]{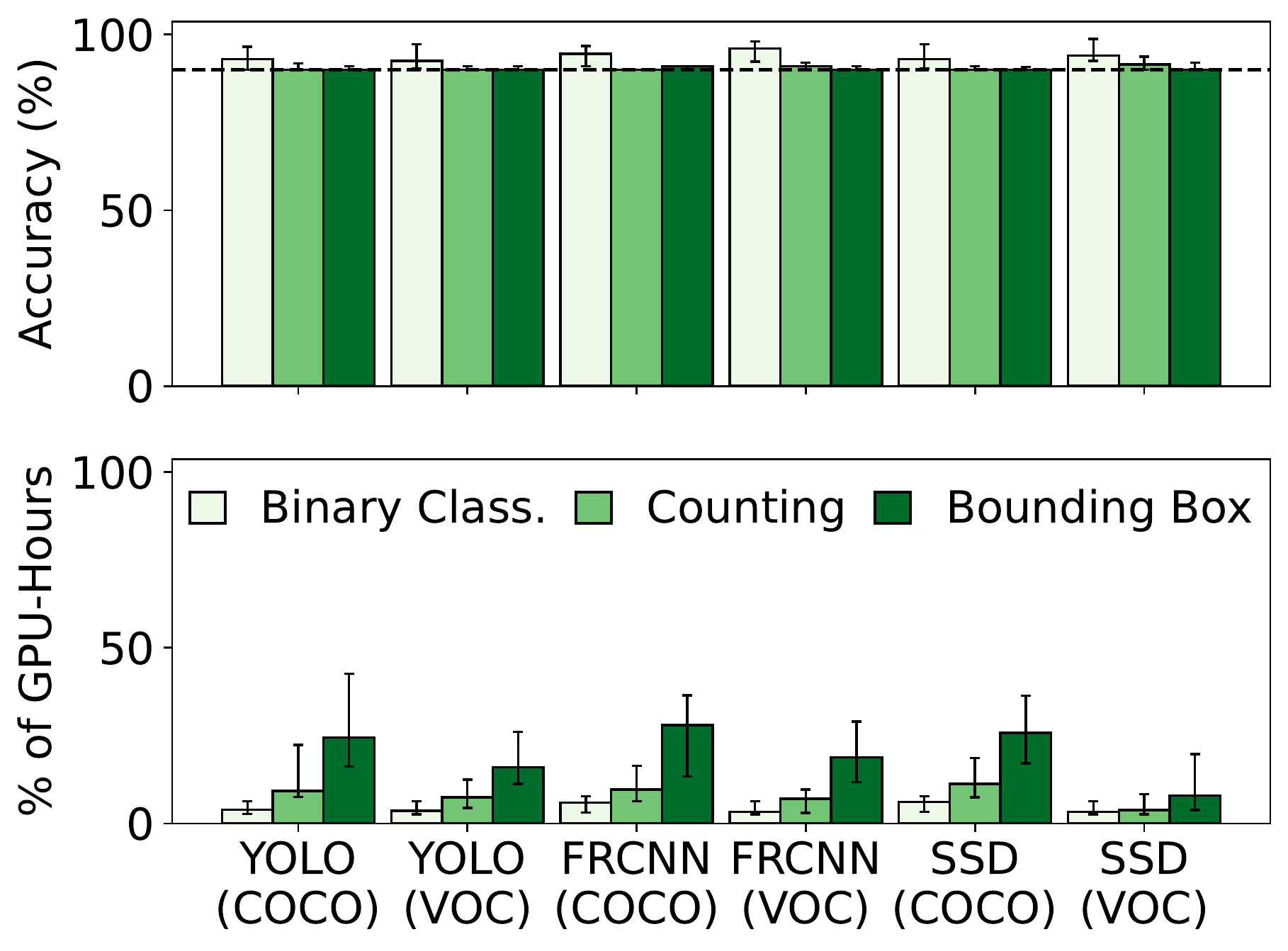}
        \label{fig: main_result_b}
        \vspace{-8pt}
        \tightcaption{{90\% Accuracy Target}}
    \end{subfigure}
     \begin{subfigure}[t]{0.33\textwidth}
        \centering
        \includegraphics[width=\columnwidth]{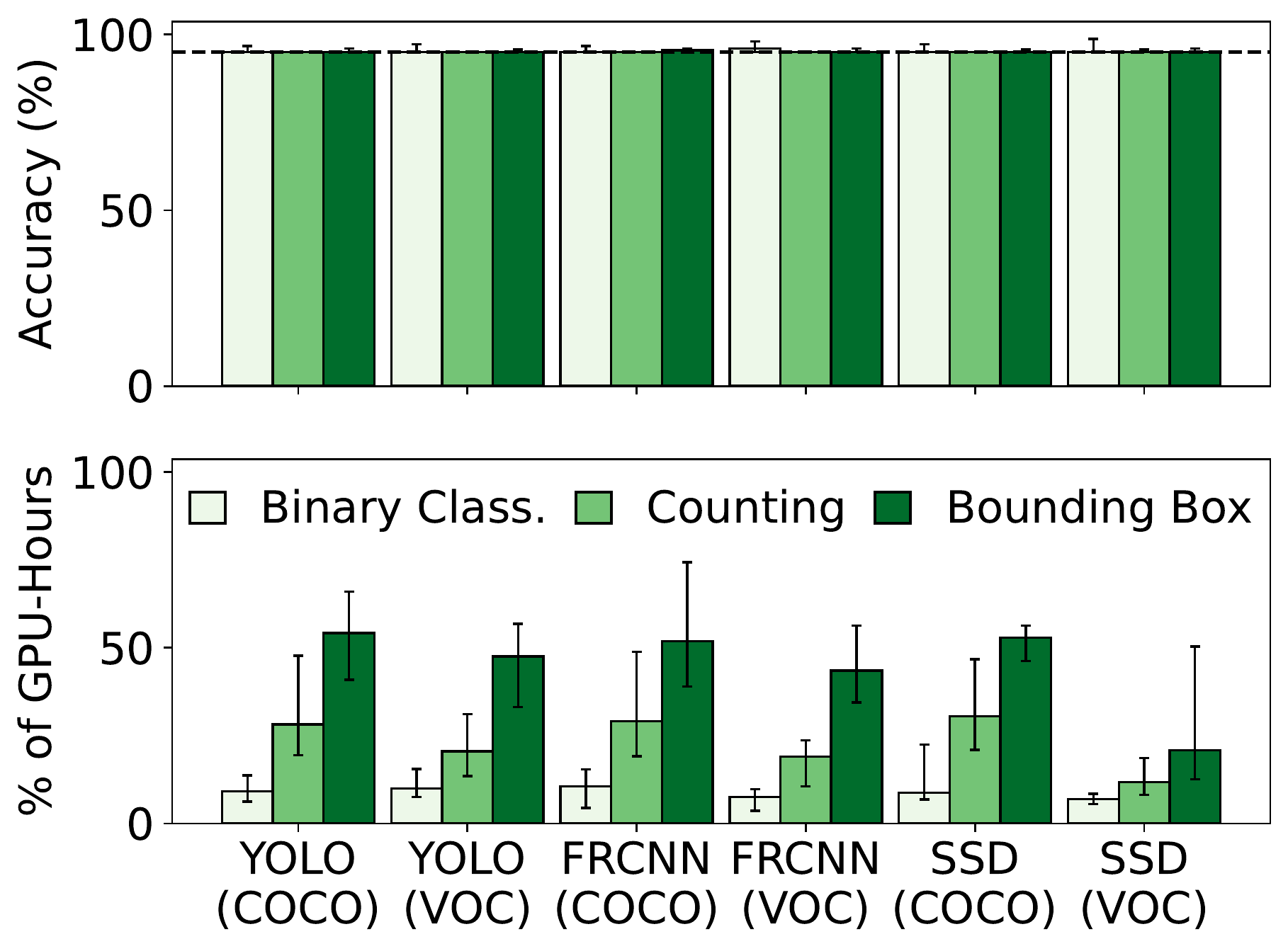}
        \label{fig: main_result_c}
        \vspace{-8pt}
        \tightcaption{{95\% Accuracy Target}}
    \end{subfigure}
    \vspace{-1pt}
    \tightcaption{\name{}'s 
    query execution performance across CNNs, query types, and accuracy targets; results are aggregated across object types. Bars summarize the distributions of per-video average result accuracy (top) and percentage of GPU-hours required to generate results relative to running the CNN on all frames (bottom). Bars list medians with error bars spanning 25-75th percentiles.}
    \vspace{6pt}
    \label{fig:main_result}
\end{figure*}
\subsection{Methodology}
\label{ss:methodology}

\begin{table}[t!]
    \setlength\arrayrulewidth{1pt}
    \footnotesize\centering
    \begin{tabular}{ccc}
    \toprule
    \textbf{Camera location} & \textbf{Resolution} \\ \midrule
    Auburn, AL (University crosswalk + intersection)~\cite{auburn} & $1920\times1080$ \\
    Atlantic City, NJ (Boardwalk)~\cite{ocean} & $1920\times1080$ \\
    Jackson Hole, WY (Crosswalk + intersection)~\cite{jacksonhole} & $1920\times1080$ \\
    Lausanne, CH (Street + sidewalk)~\cite{lausanne} & $1280\times720$ \\
    Calgary, CA (Street + sidewalk)~\cite{calgary} & $1280\times720$ \\
    South Hampton, NY (Shopping village)~\cite{hampton} & $1920\times1080$ \\
    Oxford, UK (Street + sidewalk)~\cite{oxford} & $1920\times1080$ \\
    South Hampton, NY (Traffic intersection)~\cite{southhampton} & $1920\times1080$ \\ \bottomrule
    \end{tabular}
    \vspace{1pt}
    \tightcaption{Summary of our main video dataset.}
    \label{t:dataset}
\end{table}

\para{Videos.} Table~\ref{t:dataset} summarizes the primary video sources used to evaluate \name{}. Video content across the cameras exhibits diversity in setting, resolution, and camera orientation (relative to the scene). From each camera, we scraped 12 hours of continuous video (at 30 fps) in order to capture varying levels of lighting and object densities (i.e., busyness). We consider additional videos and scene types in \S\ref{ss:deepdive}.

\para{Queries.} We consider the three query types (and their corresponding accuracy definitions) described in \S\ref{s:motivation}, i.e., binary classification, counting, and bounding box detection. For each type, we ran the query across our entire video dataset, and considered two objects of interest, people and cars, that cover drastically different size, motion, and rigidity properties; \S\ref{ss:deepdive} presents results for additional object types. We evaluated \name{} with three accuracy targets -- 80\%, 90\%, and 95\% -- and report accuracies as averages for each video. \edit{Note that accuracies are computed relative to running the model directly on all frames; as in prior systems and commercial platforms~\cite{focus-osdi18,reducto,dds,chameleon-sigcomm18,rocket}, \name{} does not aim to improve the accuracy of the provided model, and instead targets the same per-frame results at lower resource costs and delays.}

\para{CNN models.} We consider three popular architectures: (1) SSD with ResNet-50 as the base, (2) Faster RCNN with ResNet-50 as the base, and (3) YOLOv3 with Darknet53 as the base. For each, we used one version trained on the COCO dataset, and another trained on VOC Pascal. Trends for any results shown on a subset of CNNs (due to space constraints) hold for all considered models.

\para{Hardware.} Experiments used a server with an NVIDIA GTX 1080 GPU (8 GB RAM) and 18-core Intel Xeon 5220 CPU (2.20 GHz; 125 GB RAM), running Ubuntu 18.04.3.

\para{Metrics.} In addition to accuracy, we evaluate query execution performance of all considered systems (\name{}, Focus~\cite{focus-osdi18}, and NoScope~\cite{noscope-vldb17}) in terms of the number of GPU-hours required to generate results. We report GPU-hours for two reasons: (1) CNN execution (on GPUs) accounts for almost all response generation delays with all three systems, and (2) it is directly applicable to all of the systems, e.g., it incorporates NoScope's specialized CNNs. For preprocessing, we report both GPU- and CPU-hours since \name{} only requires the latter. As in prior work~\cite{focus-osdi18,noscope-vldb17}, we exclude the video decoding costs shared by all considered systems.

\subsection{Query Execution Speedups}

Figure~\ref{fig:main_result} evaluates \name{}'s query response times relative to a naive baseline that runs the CNN on all frames. \name{} always used the same, model-agnostic index per video.

There are three points to take away from Figure~\ref{fig:main_result}. First, across all of the conditions, \name{} \emph{consistently} meets the specified accuracy targets. Second, the percentage of GPU-hours required to meet each accuracy target with \name{} grows as we move from coarse classification and counting queries to finer-grained bounding box detections. For example, with a target accuracy of 90\%, the median percentage of GPU-hours across all models was \fillin{3-6}\%, \fillin{4-11}\%, and \fillin{8-28}\% for the three query types, respectively. Third, the percentage of GPU-hours also grows as the target accuracy increases for each query type. For instance, for counting queries, the percentage (across all CNNs) was \fillin{3-5}\% when the target accuracy was 80\%; this jumps to \fillin{12-30}\% when the target accuracy grows to 95\%.
The reason is intuitive: higher accuracy targets imply that \name{} must more tightly bound the duration over which results are propagated (to limit propagation errors) by running the CNN on more frames. 

\begin{table}[]
\footnotesize
\centering
\begin{tabular}{|c|c|c|c|c|}
\hline
Object Type $\rightarrow$ & \multicolumn{2}{c|}{\textbf{People}} & \multicolumn{2}{c|}{\textbf{Cars}} \\ \hline
Query Type $\downarrow$& Acc. & \% GPU-hrs & Acc. & \% GPU-hrs\\ \hline

\textbf{Binary Classif.} & 92\% & 6\% & 98\% & 3\% \\ \hline
\textbf{Counting} & 90\% & 11\% & 90\% & 7\% \\ \hline
\textbf{Bounding Box} & 91\% & 27\% & 90\% & 16\% \\ \hline
\end{tabular}
\vspace{2pt}
\tightcaption{Average accuracy and percentage of GPU-hours (relative to the naive baseline) for different query types and objects of interest. Results list median per-video values across all CNNs.}
\vspace{2pt}
\label{tab:breakdown_across_objects}
\end{table}

\para{Different object types.}
Table~\ref{tab:breakdown_across_objects} reports the results from Figure~\ref{fig:main_result} separately per object type. As shown, the high-level trends from above persist for each. However, for a given query type, the percentage of required GPU-hours is consistently lower when considering cars versus people. The reason is twofold. First, inconsistencies in CNN results are more prevalent for people since they appear as smaller objects in our scenes (\S\ref{ss:repframes}). Second, cars are inherently more rigid than people, and thus deliver more stability in the anchor ratios that \name{} relies on for bounding box propagation (\S\ref{ss:propagate}); consequently, propagation errors for bounding box queries grow more quickly with people than cars. \name{} handles both issues by running the CNN on more representative frames.

\begin{figure}[t]
    \centering
    \includegraphics[width=0.9\columnwidth]{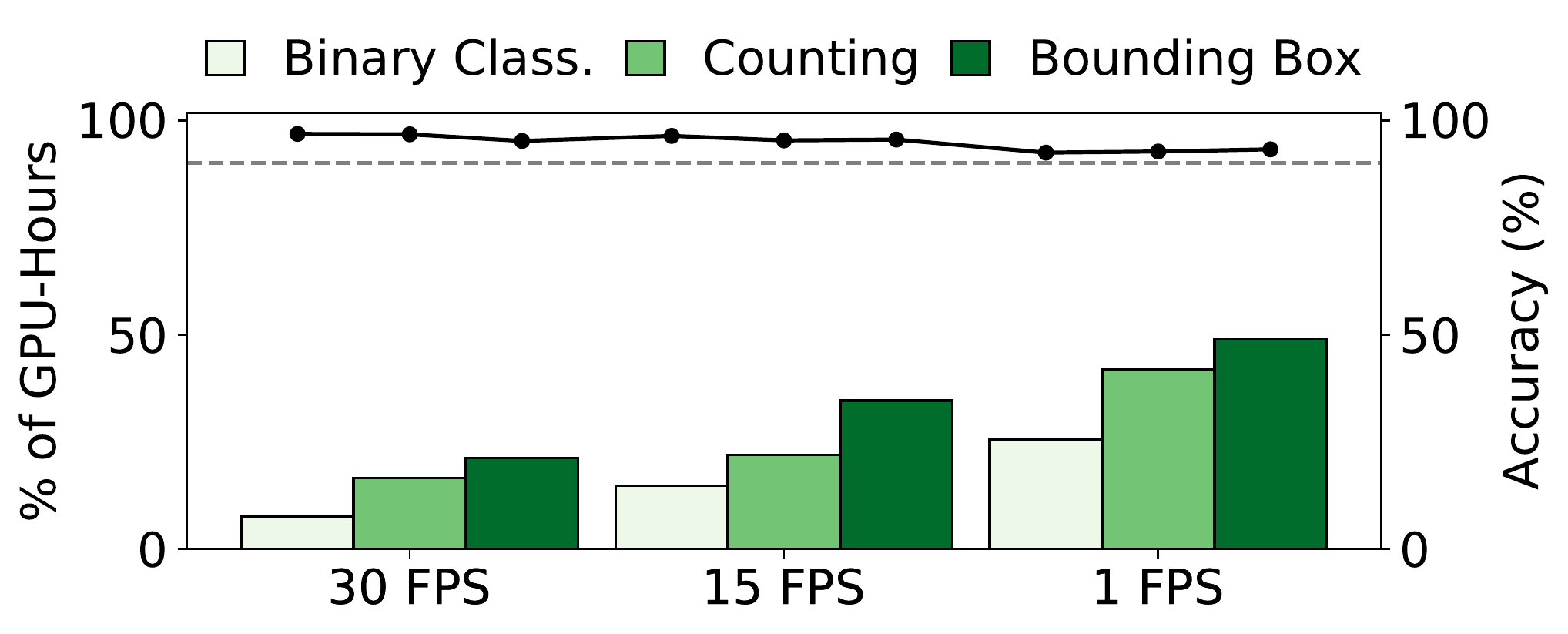}
    \tightcaption{Average accuracy (line) and percentage of GPU hours (relative to the naive baseline) for different video sampling rates. Results are listed for the median video, and consider YOLOv3+COCO and a 90\% accuracy target.} 

    \label{fig:reduced_fps_eval}
\end{figure}

\para{Downsampled video.}
Users may issue queries on sampled versions of each video~\cite{focus-osdi18}. We evaluated \name{} with three different sample rates: \{30, 15, 1\} fps. 
Although the number of considered frames drops, Figure~\ref{fig:reduced_fps_eval} shows that \name{}'s query execution speedups persist when operating over downsampled videos. For instance, with 1-fps video, \name{} requires only \fillin{25-49}\% of the GPU-hours that the naive baseline would need across all query types. Figure~\ref{fig:reduced_fps_eval} also shows that \name{}'s ability to consistently meet accuracy targets holds across sampling rates. We find that \name{} can hit accuracy targets without resorting to running the CNN on all frames because object keypoints -- the primitive that \name{} tracks across frames during both trajectory construction (preprocessing) and detection propagation (query execution) -- typically persist across frames even at these sample rates. For instance, \name{} matches \fillin{85\%} of the median object's keypoints across the 29-frame gap induced by the 1-fps rate.

\subsection{Comparison to State-of-the-Art}
\label{ss:comparisons}

\begin{figure}[t]
    \begin{subfigure}[b]{0.9\columnwidth}
        \centering
        \includegraphics[width=\columnwidth]{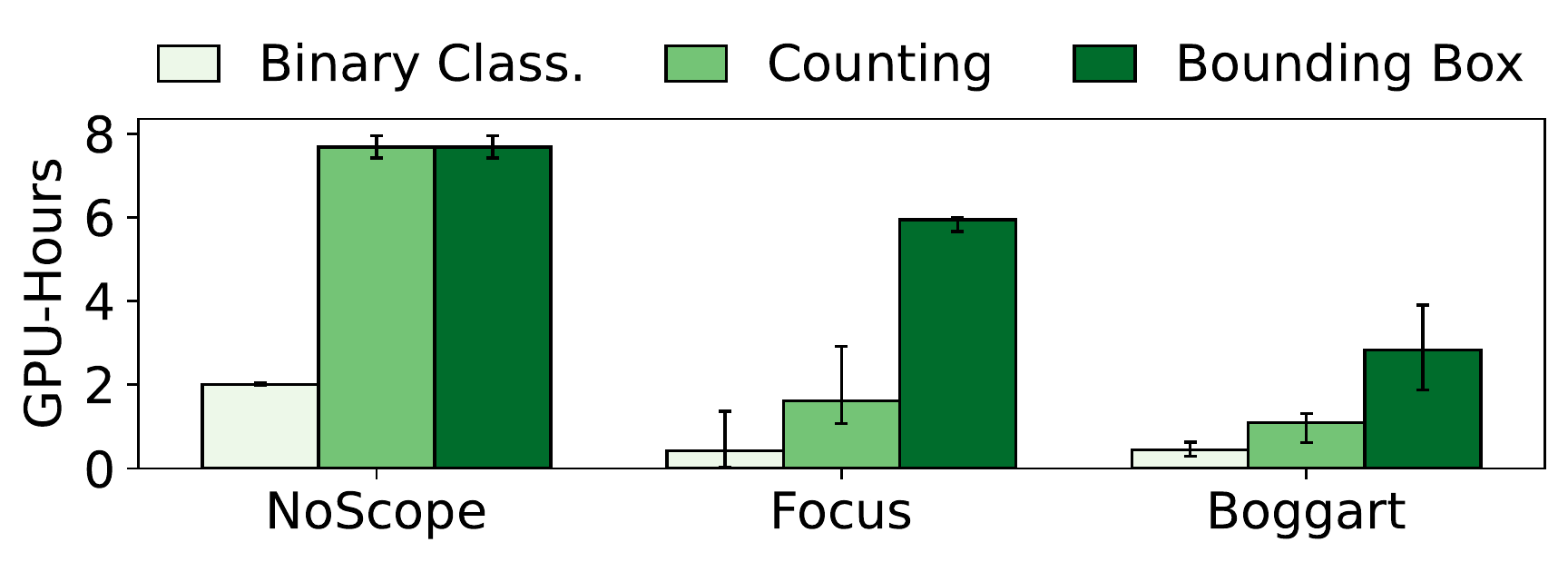}
        \vspace{-8pt}
        \tightcaption{Query execution efficiency. Bars list values for the median video, with error bars spanning 25-75th percentiles.}
        \vspace{14pt}
        \label{fig:query_comp}
    \end{subfigure}
    \begin{subfigure}[b]{0.9\columnwidth}
        \centering
        \includegraphics[width=\columnwidth]{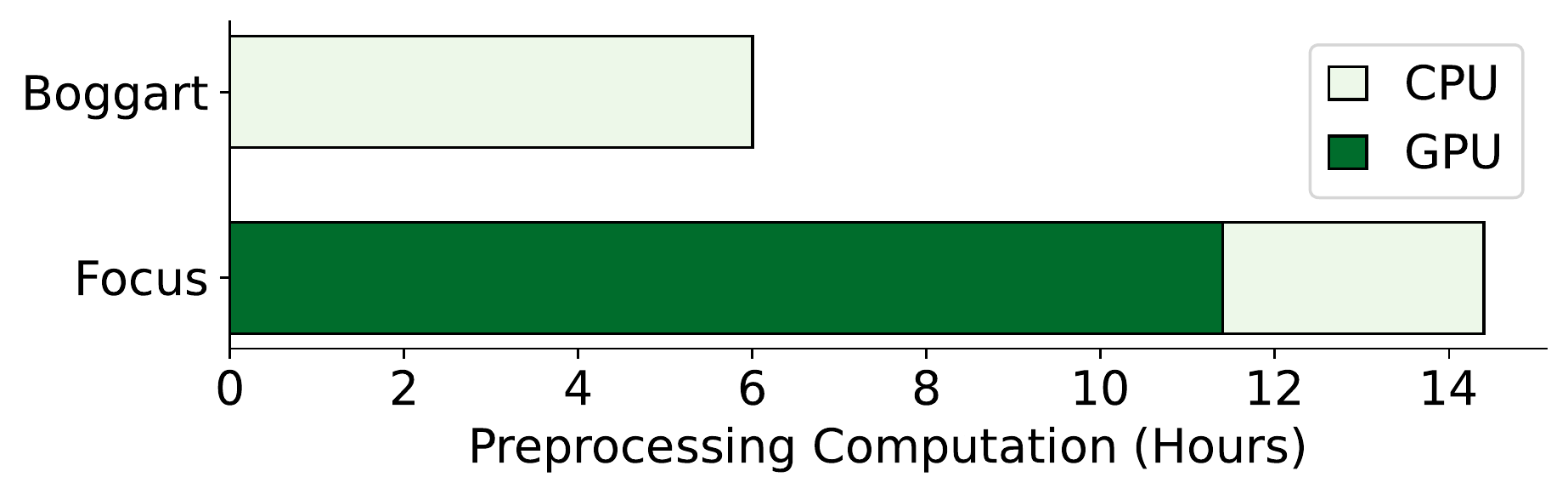}
        \vspace{-8pt}
        \tightcaption{Preprocessing efficiency. Bars list GPU/CPU-hours for the median video. Note that NoScope does not perform preprocessing.}
        \label{fig:ingest_comp}
    \end{subfigure}
    \vspace{18pt}
    \tightcaption{Comparing \name{}, Focus~\cite{focus-osdi18}, and NoScope~\cite{noscope-vldb17}. Results use YOLOv3+COCO and a target accuracy of 90\%.}
    \vspace{5pt}
    \label{fig:comparisons}
\end{figure}

We compared \name{} with two recent retrospective video analytics systems: (1) NoScope~\cite{noscope-vldb17}, which only employs optimizations during query execution, and (2) Focus~\cite{focus-osdi18}, which performs model-specific preprocessing by assuming a priori knowledge of user CNNs. \S\ref{ss:existing} details each system.

For these experiments, we set the user-provided CNN to be YOLOv3+COCO, and the accuracy target to be 90\%. Our Focus implementation used Tiny YOLO~\cite{redmon2018yolov3} as the specialized/compressed model (i.e., we ran Focus as if it knew the user CNN a priori), while NoScope used all of its open-source models. Following the training methodology used in both papers, we train the specialized/compressed models on 1-fps versions of the first half (i.e., 6 hours) of each video in our dataset, and run queries on the second half of each video. 

\para{Query Execution.} Figure~\ref{fig:query_comp} compares the query response times of all three systems. As shown, Focus requires \fillin{5}\% fewer median GPU-hours than \name{} for binary classification queries. The main reason is that Focus' model-specific preprocessing (i.e., clustering of objects) enables more result propagation than \name{}'s general, model-agnostic trajectories, i.e., Focus can propagate labels across objects, whereas \name{} can propagate labels only along a given object's trajectory (\S\ref{s:overview}). Median propagation distances for results from the full CNN are \fillin{58} and \fillin{44} frames with Focus and \name{}.

Summing Focus' classifications to generate per-frame counts was insufficient for our 90\% target. Thus, for counting queries, we performed favorable sampling until Focus hit 90\% in each video: we greedily select a set of contiguous frames with constant count errors, run the CNN on a single frame, and correct errors on the remaining ones in the set. Even with such favorable sampling, \name{} required \fillin{33\%} fewer GPU-hours than Focus for counting queries.

Bounding box detections paint a starker contrast, with \name{} needing \fillin{52}\% fewer GPU-hours than Focus. Unlike with classification labels, Focus cannot propagate bounding boxes across frames. Instead, to accelerate these queries, Focus relies on binary classification, and runs the full CNN on all frames with an object of interest (to obtain their bounding boxes); for our videos, this translates to running the full CNN on \fillin{63-100}\% of frames. In contrast, \name{} propagates bounding boxes along each trajectory (median propagation distance of \fillin{23} frames) and reduces CNN tasks accordingly.

Compared to NoScope, \name{}'s query execution tasks consume \fillin{19-97}\% fewer GPU-hours across the three query types. \name{}'s speedups here are largely due to three reasons. First, NoScope does not perform preprocessing, and instead must train and run inference with its specialized/compressed CNNs during query execution. Second, results are not propagated across frames. Third,  bounding box detections are sped up only via binary classification; note that NoScope performs binary classification on each frame (not object, like Focus), so counting queries cannot simply sum classification results per frame and instead are treated as bounding box queries.

\para{Preprocessing.}
Figure~\ref{fig:ingest_comp} shows that \name{}'s preprocessing
tasks require \fillin{58}\%
fewer computation hours than Focus' tasks. The discrepancy is driven by both the training and inference costs that Focus incurs by using a specialized/compressed model. Note that \emph{all} of \name{}'s preprocessing costs are CPU-based, while Focus' costs are dominated (\fillin{79}\%) by (more monetarily expensive) GPU operations. Further, \name{}'s preprocessing runs once per video to support all future CNNs.
To avoid accuracy drops (\S\ref{ss:limitations}), Focus would have to run preprocessing for each CNN it wishes to support, leading to higher costs and potential for wasted work.

\subsection{Profiling \name{}}
\label{ss:deepdive}

\para{Dissecting \name{}'s performance.}
\name{}'s preprocessing delays are dominated (\fillin{83}\% on the median video) by the extraction of SIFT keypoints across frames; background estimation, trajectory construction, and clustering together account for only \fillin{17}\% of the time. Query execution profiles are similar, with CNN inference on centroid chunks and representative frames contributing \fillin{7}\% and \fillin{91}\% of runtime; result propagation (mostly for detections) takes the remaining \fillin{2}\%.

\begin{figure}[t]
    \centering
    \includegraphics[width=0.85\columnwidth]{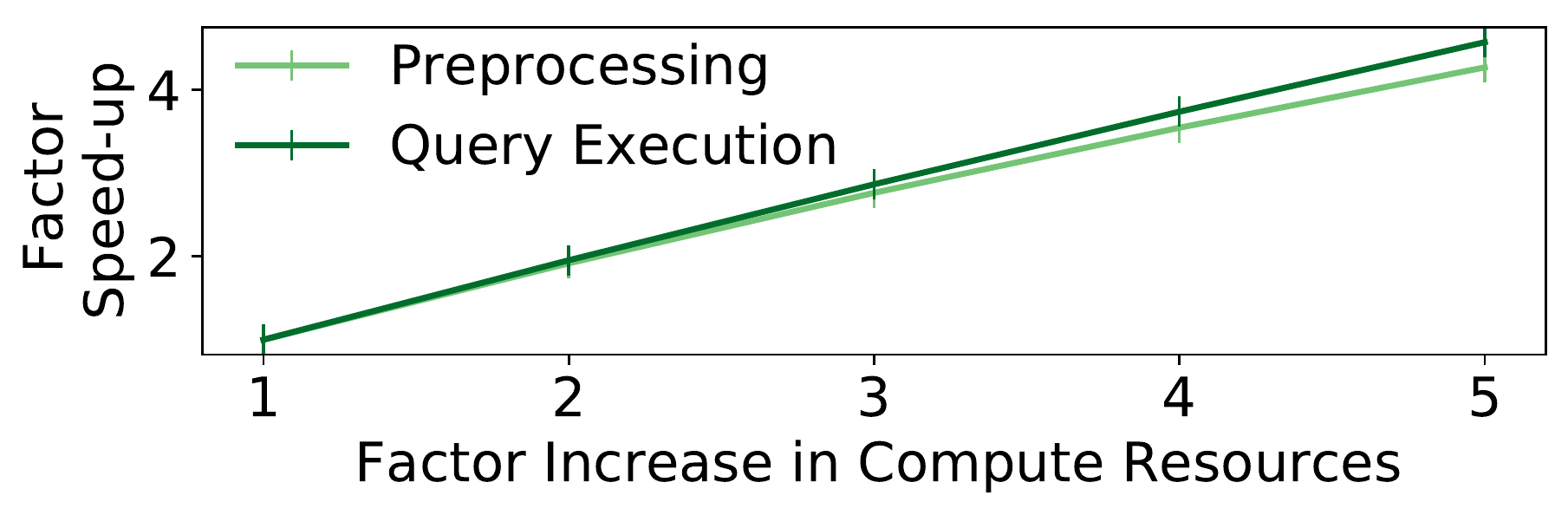}
    \tightcaption{\name{}'s performance with increasing compute resources. Resource factors are multiples of the 18-core CPU and single GPU listed in \S\ref{ss:methodology}. Results consider YOLOv3+COCO, a 90\% accuracy target, and the median video.} 
    \vspace{4pt}
    \label{fig:parallelism}
\end{figure}

\para{Resource scaling.} Figure~\ref{fig:parallelism} shows that \name{}'s preprocessing and query execution performance scale nearly linearly with increasing CPU and GPU resources, respectively. The reason is that feature extraction and CNN inference, the tasks that dominate delays in the two phases, inherently operate on a per-frame basis and can thus naturally be parallelized across frames.
Note that these results only consider parallel processing within each chunk; \name{} can also parallelize across chunks since trajectories are bound to single chunks, i.e., no cross-chunk state sharing (\S\ref{s:query}).

\para{Storage costs.} \name{}'s preprocessing generates, on average, \fillin{306 MB} of data per 1 hour of video; for context, the video is \fillin{1 GB} when encoded with H.264. Note that \fillin{98}\% of \name{}'s storage overheads are for keypoints used to propagate bounding boxes; blobs and trajectories consume only \fillin{2}\%.

\para{Sensitivity to parameters.}
\name{} includes parameters for video chunk size (default: 1 min) and target number of clusters (default: centroids cover 2\% of video). On average, we find that \name{}'s performance is largely insensitive to both: varying chunk sizes from 0.2-10 min and the videos covered by centroids from 0.5-5\% altered \name{}'s performance by less than \fillin{5}\% (note that accuracy never dropped below the targets). However, the effects of each parameter are more pronounced on short amounts of video and are dependent on the content being considered. More specifically, smaller chunk sizes reduce the potential result propagation, but also shrink cluster centroids and increase the potential for parallel processing. Similarly, more clusters implies fewer suboptimalities in the selection of representative frames, but also additional centroids on which to run the CNN.

\para{Generalizability.} To further evaluate \name{}'s ability to generalize, we ran experiments with three additional videos (3 hours each) and new object types specific to those scenes: birds in nature~\cite{birds_video}, boats in a canal~\cite{canal_video}, and \fillin{people, cups, chairs, and tables} in a restaurant~\cite{beachbar_video}. For these experiments, we ran \name{} in the same way as above, i.e., it is not tuned in any way to the video or objects of interest. We also ran experiments considering different object types (trucks and bicycles) in the traffic videos from Table~\ref{t:dataset}; these experiments used the exact same indices as in our main evaluation. All results exhibit similar trends as above, with \name{} always meeting accuracy targets (80\%, 90\%, 95\%) and running CNNs on only \fillin{11.7-34.2\%}, \fillin{11.7-53.4\%}, and \fillin{12.6-56.7\%} frames for binary classification, counting, and detection.
\section{Additional Related Work}
\label{s:related}

\para{Live video analytics.} Multiple systems accelerate queries on live video, with optimizations along the following axes: (1) profiling pipeline knobs to identify cheaper (but accurate) configurations~\cite{chameleon-sigcomm18,videostorm-nsdi17}, (2) integrating on-camera or edge server resources for partial inference, frame filtering, or reusing results from prior frames~\cite{glimpse-sensys15, reducto, filter-forward-sysml19, vigil-wireless-surveillance, video-analytics-drones, dnn-black-box-hotedgevideo19, foggycache, potluck, cachier, deepcache, marlin, euphrates}, (3) content/model-aware encoding to reduce data transfers~\cite{dds, cloudseg}, and (4) spatiotemporal coordination for efficient multi-camera queries~\cite{spatula,optasia-socc16}. These systems target an entirely different computational model (stream processing vs. ``after-the-fact'' querying) and thus face a different set of goals, optimization knobs, and constraints, e.g., by not having the entire dataset up front, live analytics can only propagate results to later frames.

\para{Accelerating GPU tasks.} One line of work optimizes DNNs for accelerated inference via distillation~\cite{hinton2015distilling}, quantization~\cite{courbariaux2016binarized, jacob2018quantization,zhu2016trained}, or pruning~\cite{liu2017learning, blalock2020state}. Another direction targets faster inference for a model, either through better scheduling of GPU resources across inference tasks~\cite{jain2018dynamic, ukidave2016mystic, nexus-sosp19}, or hardware acceleration~\cite{hard1,hard2,hard3,hard4}. These works are complementary to \name{}, which focuses on reducing the number of frames on which inference must be performed.

\para{Video Object Detection}. In addition to those in \S\ref{s:ingest}, \name{} builds on a line of work in the CV community that aims to leverage the spatiotemporal aspect of video to accelerate detection and classification tasks. These techniques swap inference on sampled frames with optical flow networks that extend results from earlier frames~\cite{zhu2017deep, zhu2018towards, chai2019patchwork, wang2019fast, kang2016object, deng2019relation, guo2019progressive, wang2018fully, he2020temporal, zhu2017flow, bertasius2018object, mao2018catdet, STLattice2018CVPR, chen2020memory}, and are thus similar in spirit to \name{}'s result propagation strategy. However, unlike \name{}, these approaches are model-specific, in that the networks used for propagation must be trained according to the specific CNN (e.g., its feature extractor) used in the target query.

\para{Video storage and indexing.} Many systems balance video storage and lookup costs for specific query types~\cite{effisearch,shotbound,patrec} or CNNs~\cite{awsdeeplens,vstore-eurosys19,smol,scanner}. \name{} is complementary to these works in that its focus is on performing generalizable preprocessing and accelerating response generation after video frames are loaded into memory.
\section{Conclusion}
\label{s:conclusion}

This paper described \name{}, a system for retrospective video analytics that supports the general ``bring your own model'' interfaces that are now commonplace in commercial platforms. To meet the core accuracy, speed, and efficiency goals of those platforms, \name{} holistically rethinks the query execution process, introducing cheap techniques to generate comprehensive (but imprecise) indices during preprocessing, and later use those indices to limit costly inference while bounding accuracy drops from imprecisions. Our results show that such generality can come at low cost, as \name{} routinely outperforms prior, model-specific approaches.

\label{lastpage}
\balance
\Urlmuskip=0mu plus 1mu\relax
\bibliographystyle{abbrv}
\bibliography{paper}

\end{sloppypar}
\end{document}